\documentclass[12pt]{article}

\usepackage[titletoc,title]{appendix}
\usepackage{afterpage}

\usepackage{picture}
\usepackage{graphicx}
\usepackage{tikz}
\usetikzlibrary{arrows}

\usepackage{amsmath}
\usepackage{amsthm}
\usepackage{amssymb}
\usepackage{upgreek}
\usepackage{isomath}
\usepackage{bbm}
\usepackage{verbatim}
\usepackage{subfigure}
\usepackage{etoolbox}
\usepackage{float}
\usepackage{algorithmic}
\usepackage[algosection,algoruled,vlined]{algorithm2e}
\usepackage{multirow}
\usepackage{rotating}
\usepackage{epstopdf}

\numberwithin{equation}{section}
\numberwithin{figure}{section}
\numberwithin{table}{section}
\makeatletter
\renewcommand{\p@subfigure}{\thefigure}
\makeatother

\newtheorem{definition}{Definition}[section]
\newtheorem{theorem}{Theorem}[section]
\newtheorem{proposition}[theorem]{Proposition}

\newcommand{\repeatable}[2]{\makeatletter \global\expandafter\def\csname repText@#1\endcsname {#2} \makeatother #2}
\newcommand{\repeatxt}[1]{\makeatletter \expandafter\csname repText@#1\endcsname \makeatother}

\newtoggle{citeInAbstract}
\toggletrue{citeInAbstract}

\newcommand{\usecrop}[2]
{
	\newlength{\cropwidth}
	\setlength{\cropwidth}{\the\textwidth}
	\addtolength{\cropwidth}{#1}
	\newlength{\cropheight}
	\setlength{\cropheight}{\the\textheight}
	\addtolength{\cropheight}{#2}
	\usepackage[width=\the\cropwidth,height=\the\cropheight,center]{crop}
}

\usepackage{totcount}

\usepackage{atbegshi}
\newif\ifRP
\newbox\RPbox
\setbox\RPbox\vbox{\vskip1pt}
\makeatletter
\AtBeginShipout{%
  \ifRP
    \AtBeginShipoutDiscard%
    \global\setbox\RPbox\vbox{\unvbox\RPbox
      \box\AtBeginShipoutBox\kern\c@page sp}%
  \fi
}%
\renewcommand{\RPtrue}{%
  \clearpage
  \ifRP\RPfalse\fi
  \global\let\ifRP\iftrue
}%
\renewcommand{\RPfalse}{%
  \clearpage
  \global\let\ifRP\iffalse
  \setbox\RPbox\vbox{\unvbox\RPbox
    \def\protect{\noexpand\protect\noexpand}%
    \@whilesw\ifdim0pt=\lastskip\fi
      {\c@page\lastkern\unkern\shipout\lastbox}%
  }%
}%
\makeatother

\DeclareMathAlphabet{\mathpzc}{OT1}{pzc}{m}{it}

\newcommand{\abs}[1]{\left | #1 \right |}

\newcommand{\Rn}[1]{{\mathbbm{R}^{#1}}}

\newcommand {\defeq}{\triangleq}

\newcommand {\myvec}[1] {{\mbox{\boldmath $#1$}}}

\newcommand{\bw}{{\bf w}}

\newcommand{\bbf}{{\bf f}}

\newcommand{\bg}{{\bf g}}

\newcommand{\bx}{{\bf x}}
\newcommand{\btx}{\widetilde{\bx}}
\newcommand{\tx}{\widetilde{x}}
\newcommand{\btheta}{{\boldsymbol{\theta}}}
\newcommand{\bpsi}{{\boldsymbol{\psi}}}

\newcommand{\tJ}{{J}}

\usepackage{upgreek}
\usepackage{isomath}
\usepackage{fullpage}

\title{Multi-View Kernel Consensus For Data Analysis}
\author{\fontsize{11.4}{14}\selectfont Moshe Salhov$^1$, Ofir Lindenbaum$^2$, Yariv Aizenbud$^4$, Avi Silberschatz$^3$, Yoel Shkolnisky$^4$, Amir Averbuch$^1$ \\
\small $^1$School of Computer Science, Tel Aviv University, Tel Aviv 69978, Israel \\
\small$^2$School of Engineering, Tel Aviv University, Tel Aviv 69978, Israel \\
\small$^3$Department of Computer Science, Yale University, New Haven,CT 06520-8285, USA \\
\small $^4$School of Mathematical Science, Tel Aviv University, Tel Aviv
69978, Israel }

\begin{document}
\maketitle
%\begin{frontmatter}

%\maketitle

\begin{abstract}
Input data is high-dimensional while the intrinsic dimension of this data maybe low. 
Data analysis methods aim to uncover the underlying low dimensional structure imposed by the low dimensional hidden parameters. In general, uncovering is achieved by utilizing distance metrics that considers the set of attributes as a single monolithic set.
However, the transformation of a low dimensional phenomena into  measurement of high dimensional observations can distort the distance metric. This distortion can affect the quality of the desired  estimated low dimensional geometric structure.   
In this paper, we propose to utilize the redundancy in the feature domain by analyzing multiple subsets of features that are called views. 
The proposed methods utilize the consensus between different
views to extract valuable geometric information that unifies
multiple views about  the intrinsic relationships among  several
different observations. This unification  enhances the information better than what a single view or a simple concatenations of views can  provide.
\end{abstract}

%\begin{keyword}
%Multi view \sep Dimensionality reduction \sep manifold learning \sep kernel PCA \sep Diffusion Maps \sep Diffusion Distance \sep Distance Preservation
%\end{keyword}

%\end{frontmatter}

\section{Introduction}

Kernel methods constitute a wide class of algorithms for
non-parametric data analysis of  high dimensional big data.
Typically, a limited set of underlying factors generates the high
dimensional observable parameters via non-linear mappings. The
non-parametric nature of these methods enables one to uncover hidden
structures in the data. These methods extend the well known
Multi-Dimensional Scaling (MDS)~\cite{cox:MDS,kruskal:MDS} method.
They are based on an affinity kernel construction that encapsulates
the relations (distances, similarities or correlations) among
multidimensional data points. Spectral analysis of this kernel
simplifies the data representation and thus simplifies its analysis.

The Principal Component Analysis
(PCA)~\cite{jolliffe:PCA,hotelling:PCA} method uses a covariance
matrix between the parameters of the analyzed data and projects the
multidimensional data points on a space that is spanned by the most
significant eigenvectors of the covariance matrix. The MDS method uses
the eigenvectors of a Gram matrix, which contains the inner-products
between the data  points in the analyzed dataset, to define a mapping
of the data points into an embedded space that preserves most of
these inner-products. PCA and MDS methods are equivalent. They represent
data points that use directions in which most of the variance in the
data is located.

 Methods such as Isomap~\cite{tenenbaum:Isomap},
LLE~\cite{LLE}, Laplacian
eigenmaps~\cite{belkin:LaplacianEigenmaps}, Hessian
eigenmaps~\cite{donoho:HessianEigenmaps} and local tangent space
alignment~\cite{yang:LTSA,zhang:LTSA} extend the MDS paradigm by
considering the manifold assumption. Under this assumption, the data
is assumed to be sampled from an intrinsic low  dimensional manifold
that captures the dependencies between observable parameters.
The corresponding spectral-based embedded spaces, which are computed by these methods identify the geometry of the manifold that incorporates the underlying factors in the data. In this paper, we focus on kernel-based spectral methods. However, kernel methods consist of many additional interesting flavors: Supervised Kernel methods~\cite{hastie01statisticallearning,Cortes1995} and Bayesian kernel methods~\cite{BayesianKM}, to name some.

Similarity assessments between members in datasets is a
crucial task for the analysis of any dataset. Important and popular kernel methods such as the above methods utilize similarity metrics that are based on Euclidean norm between features. For example, the
widely used Gaussian kernel is based on a scaled Euclidean norm between
multidimensional data points. In many cases, the given dataset
includes redundant features that relate to the underlying factors via an
unknown transformation. 

Since the unknown transformation function may have no derivative, similarity between the transformed data points can be a distorted version of the similarity between the underlying factors. Furthermore, the transformation may take place in the presence of noise, which adds additional distortion to the similarity assessment. The utilization of this distorted similarity for kernel-based data analysis may fail to uncover the desired geometry for the analysis.

In this paper, we propose two methods, which perform high dimensional data analysis, that aim to compensate for the distortions induced by the unknown transformation that computes  similarity assessment.
Both methods consider subsets of features  where each subset is defined as a view. They are used for computing similarity assessment for each view and for computing the agreement between all the computed similarities per a given data point.  Furthermore, we utilize this agreement to estimate the  corresponding inaccessible pairwise similarities between the underlying factors.

%%%%%%%%%%%%%%%%%%%%%%%%%%%%%%%%%%%%%%%%%%%%%%%%%%%%%%%%%%%%%%%%%%%%%%%%%
%ofir please edit from here
Learning from several views has motivated various
studies that have focused on classification and clustering that are
based on the spectral characteristics of  multiple datasets. Among
these studies are Bilinear Model \cite{Bilinear} and Canonical Correlation Analysis (CCA)~\cite{CCA}.
These methods are effective for clustering but neither  provides a low
dimensional geometry nor a structure for each view.  An
approach similar to CCA, which
seeks a linear transformation that maximizes the correlation among
the views, is described in~\cite{Boots}. The frameworks \cite{lai2000kernel,akaho2006kernel} extend the CCA by the application of a kernel function prior to the application of CCA. Data modeling by a bipartite graph is described in~\cite{DeSa}.  Then, based on the `minimum-disagreement' algorithm, \cite{DeSa}~partitions the
dataset. Recently, a few kernel-based methods have
proposed a model of co-regularizing kernels in both views~\cite{Kumar}. It is achieved
by searching for an orthogonal transformation, which maximizes the
diagonal terms of the kernel matrices that were obtained from all views, by
adding a penalty term that incorporates the disagreement among the
views. A mixture of Markov chains is proposed in \cite{Zhou} to
model the multiple views in order to apply spectral clustering. A
way to incorporate multiple metrics for the same data using
a cross diffusion process is described in~\cite{CDDM}. Again, the
applicability of the suggested approach is limited only for a
clustering task. Fused kernels, which  define cross diffusion process that define a cross-views diffusion distance, are proposed in~\cite{MV1,MV2}.
Furthermore, multiple low dimensional embeddings are learned from a set of views that are analyzed simultaneously in \cite{MV2}. The proposed framework in~\cite{MV1,MV2} utilizes the intrinsic relation within each view as well as the mutual relations between different views. The proposed construction is based on a cross-view model in which an implied random walk process between data points is restrained to hop between different views. Additionally, this method is applicable for clustering, classification and manifold learning.
An alternating diffusion process is proposed in \cite{lederman2014common,Lederman2015,LedermanTWLC15}. This construction is based on fusing the stochastic matrices under the assumption that a common random variable exists in the multiplied views. A Diffusion Maps (DM)\cite{coifman:DM}-based analysis of changing data is described in \cite{Coifman201479}. Additional relevant works for analyzing the geometry of the data is detailed in \cite{Berry2016439,Berry2016}.

A spectral approach for solving linear and non-linear Independent Component Analysis (ICA) problems is described in~\cite{singer:non-linearICA,singer:spectralICA}. It assumes that the data is generated by a dynamical processes in order to obtain a unique solution to a general ill-posed non-linear ICA problem. As shown in~\cite{singer:spectralICA}, a spectral approach for solving the non-linear ICA problem can be employed by using the Jacobian-based metric to construct a diffusion kernel. The eigenvectors
of the constructed kernel represent the data by its independent parameters. Several applications that use this approach are described in~\cite{Talmon2014,Kushnir2012}.

In this work, we extend the ICA framework in~\cite{singer:non-linearICA,singer:spectralICA} to perform  data analysis that their views are of either noisy or distorted. We propose to use the Jacobian-based metric to construct a diffusion kernel that incorporates similarity information that is derived from each view. This construction enables us to identify the consensus between views and compensates
for having either distortions or noise that are view specific. We consider two multi-view assumptions that depend on the given data:
\begin{enumerate}
\item The data consist of samples that are the outcome from a  dynamical
process;
\item The covariance matrix at each neighborhood of  a  multidimensional data point is accessible.
%\item The dataset $M$ has no additional assumptions;
\end{enumerate}

%%%%%%%%%%%%%%%
%In many natural and real-world applications, the measured signals are controlled by underlying
%processes or drivers. As a result, the signals are often highly structured and lie on a
%manifold. These signals exhibit highly redundant representations and their temporal evolution
%can be compactly described by a dynamical process on a low-dimensional manifold, e.g.
%[1, 2, 3, 4, 5].
%
%%%%%%%%%%%%%%

In case (1),  we assume  that the data is generated by a
dynamical process that operates on a low dimensional
manifold such as in~\cite{Kevrekidis2004,RahimiRD2005,Lin2006,Li2007,Macke2011},
to name some. This assumption enables us to compute the local
Jacobian-based distortion metric that is induced by a non-linear
transformation that maps the parameter space into an observable
space and isolates a common process across views. The uncommon dynamical processes are regarded as
interference.

%%%%%%%%%%% moshe

In case (2), we assume to have some knowledge about the data
points that allows us to compute the covariance matrix of the local neighborhood
at each data point. In this case, we utilize the
existence of the relation between the Mahalanobis distances~\cite{Mahalanobis1936} in the
extrinsic  and intrinsic
domains. This relation is utilized to define a similarity
distance that considers the inherent structure in  different
views that enable us to compensate for a view specific distortions.

%Finally, in the general case we discuss the normalization of a
%general but finite dataset. A method for estimating the width of a
%Gaussian kernel and the dimensionality of the intrinsic manifold  is
%discussed in~\cite{YoelTomo2008}. For a general dataset, we extend
%this normalization method to consider scale misalignment between
%different views. The proposed method computes the normalization of
%the given views, which maximize the approximation of the
%dimensionality of the intrinsic manifold.
%%%%%%%%%%%%%%%%%%.

The paper has the following structure.
Section~\ref{sec:problem_Formulation}  provides the problem formulation of the  multi-view embedding. Section \ref{sec:Multi-View of
Dynamical Process} details the multi-view analysis for dynamical
processes. The analysis of a dataset with an assumed accessible covariance matrix per neighborhood is
described in Section~\ref{sec:data_with_acc_cov}. Section~\ref{sec:examples} presents numerical examples.
 Section~\ref{sec:Conclusions} provides concluding remarks and future work directions.

\section{Problem Formulation}
\label{sec:problem_Formulation}
In the following, $\| \|$ denotes the standard Euclidean vector norm and $\|\|_F$
denotes the Frobenius matrix norm. Vectors are denoted by {\bf Bold} letters and vector components are denoted by a superscript $[]^r$.

Let $\mathcal{M}$ be a low-dimensional manifold that lies in the high-dimensional ambient Euclidean
 space $\mathbbm{R}^m$ and let $d \ll m$ be its intrinsic
dimension. Let $M \subseteq \mathcal{M}$ be a dataset of $\abs{M} =
n$ multidimensional data points that were sampled from $\mathcal{M}$. For data
analysis tasks, each extracted/measured feature vector $\bx_i\in M$
is assumed to have a  corresponding vector
$\btheta_i \in \Rn{d}$ of inaccessible controlling parameters whose $r^{th}$ component is $\theta_i^r$, $i=1,\ldots,n$.

Kernel methods analyze datasets such as $M$ by exploring
the geometry of the manifold  whose data points were
sampled from $\mathcal{M}$~\cite{coifman:DM,lafon:PHD,LLE,tenenbaum:Isomap}. The
computed kernel describes a measure of the data points pair-wise
similarity. A Euclidean
norm-based similarity metric between two data points $\bx_i,\bx_j \in M$ can be given by
\begin{equation}
\label{eq:kernel_def}
 K_{\varepsilon} \left( \bx_i,\bx_j\right) = h \left(
\frac{\|\bx_i - \bx_j \|^2}{\varepsilon}\right),~~ i,j=1,
\ldots, n
\end{equation}
where $\varepsilon$ is the kernel width and $h:\Rn{} \rightarrow
\Rn{}$ is a function designed such that the kernel matrix is
symmetric and positive semi-definite. However, instead of using the measured features for the similarity assessment in Eq.~\ref{eq:kernel_def},
we replace the  Euclidean norm~$\|\bx_i - \bx_j \|$ with the  Euclidean norm~$\|\btheta_i - \btheta_j\|$, $\btheta_{i},\btheta_{j}\in \Rn{d}$ that corresponds directly to the similarity between the inaccessible
corresponding controlling parameters instances.

In this paper, we propose to approximate the Euclidean distance $\|\btheta_i - \btheta_j\|$ (or the corresponding Mahalanobis distance) by utilizing the redundancy in the feature space. In order to quantify the relation between the approximated distance and the actual one, we introduce the notion of a view that is given in Definition~\ref{def:View}.
\begin{definition}[View]
	\label{def:View}
	Let $I_l$, $1 \leq l \leq \zeta$, be a subset of  $m_l$  features indices that were selected from the set of $m$ features from the  dataset $M$.
	 $M_l$ is a view of $M$ if for every $\bx_i \in M, 1 \leq i \leq n$, the vector 
	 $\btx_{i,l} \in M_l$ is a subset of $\bx_i$ that corresponds to the set of indices in $I_l$. Furthermore, two views may have overlapping features and $\zeta \leq 2^m-1$.
\end{definition}

%4343916
Under the multi-view formulation, we generate $\zeta$ views of $M$  by selecting $\zeta$ subsets of features from the features of  $M$. Given a set of $\zeta$ views, the dataset $M$ is constructed. Let  $\bx_i $  be the  concatenation (while overlapping feature are removed) of the $\zeta$ views such
that $\bx_i = \cup_{l=1}^\zeta \btx_{i,l}$ where $ \btx_{i,l} \triangleq
\left[x_{i,l}^1,\dots,x_{i,l}^{m_l} \right]$, $1 \leq l \leq \zeta $. Hence, for a given set of $\zeta$ views, the dataset $M$ is defined as a concatenation of all the views, and for a given dataset $M$ any number of the $2^m-1$ possible subsets can be chosen as views.

Additionally, we assume that the $l$th view is the outcome of the
function $\bbf_l:\Rn{d} \times \Rn{k_l} \to \Rn{m_l}$  such that $\btx_{i,l} = \bbf_l \left(
\btheta_i, \bpsi_{i,l} \right)$, where $\bpsi_{i,l} \in \Rn{k_l} $ is the $k_l$-dimensional view-specific controlling parameters such that $k_l \leq k_{max} <\infty$.

% bi-Lipschitz

Our goal in this work is to find a function
$G:\Rn{\zeta} \rightarrow [0,1]$ such that the desired kernel
similarity $K_{\varepsilon} \left(  \btheta_i, \btheta_j \right)$, $i,j = 1,...,n$, is approximated  by
\begin{equation}
\label{eq:multi-view_kernel} K_{\varepsilon} \left(  \btheta_i,
\btheta_j  \right) \approx G \left(K_{\varepsilon_1}
\left(\btx_{i,1},\btx_{j,1} \right), \dots ,K_{\varepsilon_l}
\left(\btx_{i,l},\btx_{j,l} \right), \dots , K_{\varepsilon_{\zeta}}
\left(\btx_{i,\zeta},\btx_{j,\zeta} \right) \right)
\end{equation}
where $\varepsilon_{l}$, $1 \leq l \leq \zeta$, is adapted to the characteristics of the $l$th view and $K_{\varepsilon}$ is defined in Eq.~\ref{eq:kernel_def}. We call this kernel (Eq.~\ref{eq:multi-view_kernel}) the multi-view kernel.

We aim to find the underlying intrinsic
geometry of $\mathcal{M}$.  By identifying the consensus between the
$\zeta$ views, we are able to approximate a kernel that is related to
the inaccessible controlling parameters $\btheta_i$.

\section{Multi-View of Dynamical Process}
\label{sec:Multi-View of Dynamical Process} In this section, we
analyze multi-views that were generated by a dynamical system. The following
analysis  extends the work in~\cite{singer:non-linearICA} and adapts the generic state-space formalism
in~\cite{singer:non-linearICA,Talmon2014} to a
variety of applications. We assume that the data points in $M$ are
the outputs from non-linear functions of independent stochastic
It\^{o} processes. Assume that $\btheta_i$ are samples from the It\^{o} process $\btheta$ such that $\btheta_i \defeq \btheta \left[t_i \right],~1 \leq i \leq n$, where $t_i$ is a time instance. The dynamics of this process
is described by  the normalized stochastic differential equations of the form
\begin{eqnarray}
\label{eq:dynimic model} d \theta^r = a^r \left( \theta^r
\right) dt + dw^r,~~~ r=1,\dots,d
\end{eqnarray}
where for all $\vartheta \in \Rn{}$, $0 \leq a\left(\vartheta \right)^r \leq a_{max} < \infty$ are the unknown drift coefficients that are assumed to be Riemann integrable and $w^r$ are the independent
white noises. For simplicity, we
consider here processes that are normalized to have a unit variance
noises.

The $d$-dimensional vector $\btheta_i$ is inaccessible. Its non-linear noisy mapping is
\begin{equation}
\label{eq:noisy_dynimic model} \btx_{i,l} = \bbf_l
\left(\btheta_i,\bpsi_{i,l}\right), ~~~  l = 1, \ldots, \zeta, \quad i = 1, \ldots, n,
\end{equation}
where  $\bbf_l$ is assumed to be differentiable,  bi-Lipschitz  and $\bpsi_{i,l} \defeq \bpsi _l \left[t_i \right]$ is a $k_l$-dimensional sample from an  It\^{o} process given
by
\begin{eqnarray}
\label{eq:dynimic noise_model} d \psi_{l}^r = a^{d+r} \left(
\psi_{l}^r \right) dt + dw_{l}^{d+r},~~~ r=1,\ldots,k_l, ~~l = 1, \ldots, \zeta,
\end{eqnarray} 
where the unknown drift coefficients are assumed to be Riemann integrable and for all $\vartheta \in \Rn{}$ it is bounded as 
\begin{equation}
\label{eq:a_max}
 0 \leq a\left(\vartheta \right)^{{d+r}} \leq  a_{max} < \infty,~~~ r=1,\ldots,k_l, ~~l = 1, \ldots, \zeta.
\end{equation}
Additionally, $w^{d+r}$ are the independent white noises. We consider $\bpsi_{l}$ as an underlying interference process that corrupts the $l$th view in the sense of Eq.~\ref{eq:noisy_dynimic model}. This interference distorts the pair-wise similarities. Proposition~\ref{cor:interference_dense} quantifies the probability of each independent process $\psi_{l}^r$ to return to its starting point after time $T < \infty$.
\begin{proposition}
\label{cor:interference_dense}
Let $\psi_{i,l}^r$ and $\psi_{j,l}^r$ be the two $r$th elements of the $k_l$-dimensional sample from  $\bpsi_{l}$ with Riemann integrable drift coefficients $a^{r}$. Without-loss-of-generality, assume $i>j$ and since the dataset acquisition process is assumed to have a finite duration we have $t_i-t_j = T \leq T_{max} < \infty$. Then, for every $\varepsilon_1>0$, we have $P \left( |\psi_{i,l}^r - \psi_{j,l}^r| \leq \varepsilon_1 \right) > 0$.
\begin{proof}
From the definition of the It\^{o} process we have 
\begin{equation}
\label{eq:psi_def}
\psi_{i,l}^r - \psi_{j,l}^r =  \int_{t_j}^{t_i} a^{d+r} \left(
\psi_{l}^r \right) dt + \int_{t_j}^{t_i}  dw_{l}^{d+r},
\end{equation}
where the first integral is a Riemannian integral and the second is an It\^{o} integral. Let $B_a$ be the result of the first integral given by 
\begin{equation}
\label{eq:B_a_def}
B_a\left(r,l\right)  = \int_{t_j}^{t_i} a^{d+r} \left(
\psi_{l}^r \right) dt.
\end{equation}
Since  $T < \infty$ is the time difference between two realized time stamps and from Eq.~\ref{eq:a_max}, $a^{d+r}$ is assumed to be bounded by $a_{max}$ and further assumed to be Riemann integrable then there exists a bound  
\begin{equation}
B_{max}  = \int_{t_j}^{t_j+T_{max}} a_{max} dt,
\end{equation}
such that for $r=1,\ldots,k_l$ and $l = 1, \ldots, \zeta$. the integral from Eq.~\ref{eq:B_a_def}
is bounded by $B_a\left(r,l\right) \leq  B_{max}<\infty$. 

The It\^{o} integral is a random variable with $ Z \triangleq \int_{t_j}^{t_i}  dw_{l}^{d+r} \sim \mathcal{N}(0,T)$. Hence, for every  $\varepsilon_1 > 0$ there exists
$\varepsilon_2 > 0$ for which the probability of $\{ Z: |Z| \leq |B_{max}+\varepsilon_1| \}$ is larger than $\varepsilon^2_2$. Furthermore, from Eq.~\ref{eq:a_max} we have
\begin{equation}
P \left( |B_a\left(r,l\right) - Z| \leq \varepsilon_1 \right) \geq P \left( |B_{max} - Z| \leq \varepsilon_1 \right) \geq \varepsilon_2^2 > 0.
\end{equation} 
From the definition of $Z$ and from Eqs.~\ref{eq:psi_def} and~\ref{eq:B_a_def}, we have $\psi_{i,l}^r - \psi_{j,l}^r = Z + B_a $.  Since $ Z \sim \mathcal{N}(0,T)$, then we have $P \left( |\psi_{i,l}^r - \psi_{j,l}^r| \leq \varepsilon_1 \right)  = P \left( |B_a + Z| \leq \varepsilon_1 \right) \geq \varepsilon_2^2 > 0$ from the symmetry of the density of $Z$ around the origin. 
\end{proof}
\end{proposition}

From the multi-view perspective, the intrinsic controlling parameters $\btheta_i$ are common to all views, while the intrinsic controlling parameters  $\bpsi_{i,l}$ are specific to the $l$th view. We call $\btheta_i$ the intrinsic parameters of the consensus of the $i$th data point.
%We further assume that the given dataset is sufficiently rich in the sense that for all $1 \leq i,j \leq n$,  $Pr \left( \| \bpsi_{i,l} - \bpsi_{j,l}\| = 0 \right) > 0$. This assumption has the following justification.
%In the case that $Pr \left( \| \bpsi_{i,l} - \bpsi_{j,l}\| = 0 \right) = 0$, each noisy process instance $\bpsi_{i,l}$ is unique and singular since every other instance $\bpsi_{j \neq i,l} $, is not similar to it. This is the case where the data points are spread sparsely over the manifold in the high-dimensional ambient space.

We assume that the data points in $M$ reside on several patches located on a low dimensional underlying manifold in the ambient space. On the other hand, if the data is spread sparsely over the manifold, then the application of an affinity-based  kernel to the data will not reveal any patches/clusters. In this case, the data is too sparse to represent or to identify the underlying manifold structure. The available processing tools are variations of nearest-neighbor type algorithms. Therefore, data points on a low-dimensional manifold in a high-dimensional ambient space can either reside in locally-defined patches and then the methods in this paper are applicable to it, or scattered sparsely all over the manifold and thus there is no detectable coherent physical phenomenon that can provide an underlying explanation for it.

Given multi-view measurements, the dynamics of the consensus process
$\btheta$   can be identified and its underlying
geometry is revealed as described next. Each view $\btx_{i,l} $ satisfies the stochastic dynamics given by the
It\^{o} Lemma for $ i = 1,...,n$, $\quad 1\leq r \leq m_l$ and $l \leq \zeta$ such that
\begin{eqnarray}
\label{eq:ito_dynamics}
dx_{i,l}^r = & \sum_{k=1}^d \left[ \left( \frac{1}{2}
\frac{ \partial^2 f_l^r \left( \btheta_i,\bpsi_{i,l}\right)}{ \partial  \theta^k  \partial \theta^k
} + a^k \frac{ \partial f_l^r  \left(\btheta_i,\bpsi_{i,l} \right)}{ \partial  \theta^k  } \right)dt   +
 \frac{ \partial f_l^r  \left( \btheta_i,\bpsi_{i,l} \right)}{ \partial  \theta^k  } dw^{k} \right]+  \nonumber \\
& \sum_{k=1}^{k_l} \left[ \left( \frac{1}{2}
\frac{ \partial^2 f_l^r  \left( \btheta_i,\bpsi_{i,l} \right)}{ \partial  \psi_l^k  \partial \psi_l^k
} + a^{d+k} \frac{ \partial f_l^r  \left(\btheta_i,\bpsi_{i,l}\right)}{ \partial  \psi_l^k  } \right)dt +
\frac{ \partial f_l^r  \left(\btheta_i,\bpsi_{i,l}\right)}{ \partial  \psi_{l}^k  } dw^{d+k} \right].
\end{eqnarray}

The accessible $\left(r,k \right)$th elements of the  $m_l \times m_l$ covariance between the samples of the $l$th view $[C_{\btx_{i,l}}]_{r,k} = \left[ E_{d \bw} \left[ \left( d\btx_{i,l}\right) \left(d\btx_{i,l} \right)^T \right]  \right]_{r,k}$, is given for $\quad 1\leq k,r \leq m_l$ by
\begin{eqnarray}
\label{eq:ito_corr}
[C_{\btx_{i,l}}]_{r,k} =
\sum_{q=1}^d \frac{ \partial f_l^r \left(\btheta_i,\bpsi_{i,l} \right)}{ \partial  \theta^q  } \frac{
\partial f_l^k \left(\btheta_i,\bpsi_{i,l}\right)}{ \partial  \theta^q  }+\sum_{q=1}^{k_l} \frac{
\partial f_l^{r} \left(\btheta_i,\bpsi_{i,l}\right)}{ \partial  \psi_{l}^{q}  } \frac{ \partial f_l^{k} \left(\btheta_i,\bpsi_{i,l} \right)}{
\partial  \psi_{l}^q  }.
\end{eqnarray}
We utilize the fact that in this case $ E \left[  dw^k  dw^r \right] = 0$ for $k
\neq r$ since $dw^{r}$ is independent of $dw^{k}$.  The covariance matrix can be reformulated in a
matrix form as a function of the corresponding Jacobian $J_{\btx_{i,l}}$ of  $\bbf_l \left(\btheta_i, \bpsi_{i,l}\right)$  by
\begin{eqnarray}
\label{eq:ito_corr_jac}
C_{\btx_{i,l}}  = J_{\btx_{i,l}} J_{\btx_{i,l}}^T,~i=1,\ldots,n,~l=1,\ldots,\zeta.
\end{eqnarray}
\begin{proposition}
\label{lem:mhana_eucl}
Let $\btx_{i,l}$ be a noisy measurement of an It\^{o} process according to Eq.~\ref{eq:noisy_dynimic model}.
Let $\btx_{i,l}$ and $\btx_{j,l}$ be two data points from the $l$th view and let $\delta>0$ be a given threshold. Furthermore, assume that the interference in the $l$th view $\bpsi_{i,l}$ is independent of both the interference of any other view and of $\btheta_i$. Then, as the number of views $\zeta$ grows, the minimal Mahalanobis distance (over the entire set of views) approaches the Euclidean distance between  the corresponding governing parameters such that
\begin{eqnarray}
\label{eq:jac_cov}
\lim_{\zeta \rightarrow \infty } \min_{1 \leq l \leq \zeta}  \frac{1}{2} \left(  \btx_{i,l} -  \btx_{j,l} \right)^T  \Lambda_{i,j,l}^{-1} \left(  \btx_{i,l} -  \btx_{j,l} \right) = \| \btheta_{i}  - \btheta_{j} \|^2 + \phi,
\end{eqnarray}
   where $\phi =O \left( \|  \btx_{i,l} -  \btx_{j,l} \|^4  \right) $, $ \Lambda_{i,j,l}^{-1}$ is given by
\begin{eqnarray}
\label{eq:ica_lim}
\Lambda_{i,j,l}^{-1} \triangleq  C_{\btx_{i,l}}^{-1} + C_{\btx_{j,l}}^{-1},
\end{eqnarray}
and $C_{\btx_{i,l}}$ and $ C_{\btx_{j,l}}$ are the covariances
matrices that correspond to the $l$th view from data points $\btx_{i,l}$ and
$\btx_{j,l}$, respectively.
 \begin{proof}
From the assumption that $\bbf_l$  is bi-Lipschitz, $\bbf_l$ has an inverse function $\bg_l:\Rn{m_l} \to \Rn{d} \times \Rn{k_l} $  such that $\left[\btheta_i,\bpsi_{i,l} \right] = \bg_l \left(\btx_{i,l}  \right)$. The relation between the Jacobian and the covariance matrix in Eq.~\ref{eq:ito_corr_jac} is proposed in~\cite{singer:non-linearICA} to approximate the distance between the governing parameters using a single $l$-th view under the assumption that the noise is not present. We reformulate this relation to include a noise process. Expanding the function $\bg_l \left(\nu_l \right)$ in Taylor series at the point $\nu_l = \btx_{j,l}$ provides for $ 1 \leq r \leq d$,
\begin{equation}
\label{eq:taylor_1}
\theta^r_i  - \theta^r_j = \sum_{k=1}^{m_l}     \frac{\partial g_l^r \left(\btx_{j,l}\right)}{\partial \nu_l^k} \left(\btx_{i,l}^k - \btx_{j,l}^k \right) + \sum_{k=1,q=1}^{m_l}     \frac{\partial^2 g_l^r \left(\btx_{j,l}\right)}{\partial \nu_l^k \partial \nu_l^q} \left(\btx_{i,l}^k - \btx_{j,l}^k \right)\left(\btx_{i,l}^q - \btx_{j,l}^q \right) +\varphi,
\end{equation}
where $\varphi = O\left( \|\left(\btx_{i,l}- \btx_{j,l} \right) \|^3 \right)$. Similarly, the noise process elements where $1 \leq r \leq k_l$ are
\begin{equation}
\label{eq:taylor_pasi_1}
\begin{array}{ll}
\psi^r_{i,l}  - \psi^r_{j,l} & = \sum_{k=1}^{m_l}     \frac{\partial g_l^{r+d} \left(\btx_{j,l}\right)}{\partial \nu_l^k} \left(\btx_{i,l}^k - \btx_{j,l}^k \right) \\
& + \sum_{k=1,q=1}^{m_l}     \frac{\partial^2 g_l^{r+d} \left(\btx_{j,l}\right)}{\partial \nu_l^k \partial \nu_l^q} \left(\btx_{i,l}^k - \btx_{j,l}^k \right)\left(\btx_{i,l}^q - \btx_{j,l}^q \right) +\varphi .
\end{array}
\end{equation}
Expanding the function $\bg_l \left(\nu_l \right)$ in Taylor series at the point $\nu_l = \btx_{i,l}$ gives for $ 1 \leq r \leq d$,
\begin{equation}
\label{eq:taylor_2}
\theta^r_j  - \theta^r_i = \sum_{k=1}^{m_l}     \frac{\partial g_l^r \left(\btx_{i,l}\right)}{\partial \nu_l^k} \left(\btx_{j,l}^k - \btx_{i,l}^k \right) + \sum_{k=1,q=1}^{m_l}     \frac{\partial^2 g_l^r \left(\btx_{i,l}\right)}{\partial \nu_l^k \partial \nu_l^q} \left(\btx_{j,l}^k - \btx_{i,l}^k \right)\left(\btx_{j,l}^q - \btx_{i,l}^q \right) +\varphi.
\end{equation}
Similarly, the noise process elements at the point $\nu_l = \btx_{i,l}$ gives for, $1 \leq r \leq k_l$
\begin{equation}
\label{eq:taylor_pasi_2}
\begin{array}{ll}
\psi^r_{j,l}  - \psi^r_{i,l} & = \sum_{k=1}^{m_l}     \frac{\partial g_l^{r+d} \left(\btx_{i,l}\right)}{\partial \nu_l^k} \left(\btx_{j,l}^k - \btx_{i,l}^k \right) \\
& + \sum_{k=1,q=1}^{m_l}     \frac{\partial^2 g_l^{r+d} \left(\btx_{i,l}\right)}{\partial \nu_l^k \partial \nu_l^q} \left(\btx_{j,l}^k - \btx_{i,l}^k \right)\left(\btx_{j,l}^q - \btx_{i,l}^q \right) +\varphi.
\end{array}
\end{equation}
By using Eqs.~\ref{eq:taylor_1} and  \ref{eq:taylor_2} to compute $\| \btheta_i  - \btheta_j \|$ and by averaging between the two results to remove the third order terms we get
\begin{eqnarray}
\label{eq:taylor_norm_1}
\| \btheta_i  - \btheta_j \|^2 =  \frac{1}{2} \left(\btx_{i,l} -  \btx_{j,l}
\right)^T  \left( \tJ_{\btheta_i,l} \tJ_{\btheta_i,l}^T +  \tJ_{\btheta_j,l} \tJ_{\btheta_j,l}^T \right) \left(  \btx_{i,l} -  \btx_{j,l} \right) +  \phi,
\end{eqnarray}
where $ \tJ_{\btheta_i,l}$ is a $m_l \times d$ matrix that holds the partial derivative of $\bg_l \left( \nu_l \right)$ and $\left [\tJ_{\btheta_i,l} \right]_{k,r} = \frac{\partial g^r_l \left( x_{i,l} \right) }{\partial \nu_l^k}$, $1 \leq r \leq d$.
Furthermore, by using Eqs.~\ref{eq:taylor_pasi_1} and  \ref{eq:taylor_pasi_2} to compute $\| \bpsi_{i,l}  - \bpsi_{j,l} \|$ and by averaging between the two results such that all the third order terms are removed yields,
\begin{eqnarray}
\label{eq:taylor_norm_2}
\| \bpsi_{i,l}  - \bpsi_{j,l} \|^2 =  \frac{1}{2} \left(\btx_{i,l} -  \btx_{j,l}
\right)^T  \left( \tJ_{\bpsi_{i,l}} \tJ_{\bpsi_{i,l}}^T +  \tJ_{\bpsi_{j,l}} \tJ_{\bpsi_{j,l}} ^T\right) \left(  \btx_{i,l} -  \btx_{j,l} \right) + \phi,
\end{eqnarray}
where $ \tJ_{\bpsi_{i,l}}$ is a $m_l \times k_l$ matrix that holds the partial derivative of $\bg_l \left( \nu_l \right)$ where $\left [\tJ_{\bpsi_{i,l}} \right]_{k,r} = \frac{\partial g^{d+r}_l \left( x_{i,l} \right) }{\partial \nu_l^k}$, $1 \leq r \leq k_l$.
Combining Eqs. \ref{eq:taylor_norm_1} and \ref{eq:taylor_norm_2} gives
\begin{eqnarray}
\label{eq:combined_norm_1}
\begin{array}{ll}
\| \bpsi_{i,l}  ~-& \bpsi_{j,l} \|^2 + \| \btheta_i  - \btheta_j \|^2_2   = \phi  \\
& +\frac{1}{2} \left(\btx_{i,l} -  \btx_{j,l} \right)^T  \left( \tJ_{\bpsi_{i,l}} \tJ_{\bpsi_{i,l}}^T +  \tJ_{\bpsi_{j,l}} \tJ_{\bpsi_{j,l}} ^T   +  \tJ_{\btheta_i,l} \tJ_{\btheta_i,l}^T +  \tJ_{\btheta_j,l} \tJ_{\btheta_j,l}^T \right) \left(  \btx_{i,l} -  \btx_{j,l} \right).  
\end{array}
\end{eqnarray}
From Eq.~\ref{eq:jac_cov}, the definitions of $ \tJ_{l,\btheta_i}$ and $ \tJ_{l,\bpsi_i}$ and by using the bi-Lipschitz assumption for the function  $\bbf_l$ we get
\begin{eqnarray}
\label{eq:ica_dist} \frac{1}{2} \left(\btx_{i,l} -  \btx_{j,l}
\right)^T   \Lambda_{i,j,l}^{-1} \left(  \btx_{i,l} -  \btx_{j,l} \right)  +  O\left( \|\left(\btx_{i,l}- \btx_{j,l} \right) \|^4 \right) =  \|\btheta_{i}  - \btheta_{j} \|^2+ \|  \bpsi_{i,l}  - \bpsi_{j,l} \|^2.
\end{eqnarray}
The noise processes $ \bpsi_{i,l}$ and $\bpsi_{j,l} $ are independent and contribute a non-negative distortion to the Mahalanobis distance In Eq.~\ref{eq:jac_cov}.

Furthermore, from Proposition~\ref{cor:interference_dense}, we have $P \left( |\psi_{i,l}^r - \psi_{j,l}^r| \leq \varepsilon_1 \right) \geq \varepsilon^2_2 > 0$ for any $1\leq r \leq k_l$. Since $\psi_{i,l}^{r_1}$ is independent of $\psi_{i,l}^{r_2  \neq r_1}$, hence, $P \left( \|  \bpsi_{i,l}  - \bpsi_{j,l} \| < \varepsilon_3 \right) \ge \Pi_{r=1}^{k_l} P \left( |\psi_{i,l}^r - \psi_{j,l}^r|<\varepsilon_3 \right) $  for every $\varepsilon_3>0$. From Proposition~\ref{cor:interference_dense}, there exists $\varepsilon_2 $ such that
$\Pi_{r=1}^{k_l} P \left( |\psi_{i,l}^r - \psi_{j,l}^r|  \leq  \frac{\varepsilon_3}{\sqrt k_l} \right) \geq \Pi_{r=1}^{k_l} P \left( |\psi_{i,l}^r - \psi_{j,l}^r|  \leq  \frac{\varepsilon_3}{\sqrt k_{max}} \right)  \geq \varepsilon_2^{2k_{max}}>0$. Note that $ \varepsilon_2^{2k_{l}} \geq \varepsilon_2^{2k_{max}} >0$, where $\varepsilon_4 = \varepsilon_2^{2k_{max}}$ is independent of $l$ and $r$. 

Hence, as the number of independent views $\zeta$ grows, the probability for not finding a view where $ \|  \bpsi_{i,l}  - \bpsi_{j,l} \| \leq \varepsilon_3  $  diminishes as
\begin{equation}
\lim_{\zeta \rightarrow \infty }\Pi_{1 \leq l \leq \zeta }\left( 1- P \left( \|  \bpsi_{i,l}  - \bpsi_{j,l} \| \leq \varepsilon_3  \right) \right) = 0,
\end{equation}
since $\Pi_{1 \leq l \leq \zeta }\left( 1- P \left( \|  \bpsi_{i,l}  - \bpsi_{j,l} \| \leq \varepsilon_3  \right) \right) \leq \left( 1- \varepsilon_4 \right)^{\zeta} $. 
Furthermore, $\varepsilon_4$ is independent of $\zeta$ and $1 - \varepsilon_4 < 1$. Thus, for every $\varepsilon_5>0$ we have $\zeta>0$ such that,
\begin{equation}
P\left( \min_{1 \leq l \leq \zeta}   \| \bpsi_{i,l}  - \bpsi_{j,l} \|  < \varepsilon_3\right) \geq 1-\left(1-  \varepsilon_4 \right)^{\zeta}>1-\varepsilon_5.
\end{equation}

\end{proof}
\end{proposition}
Proposition~\ref{lem:mhana_eucl} suggests that similarity between all the inaccessible parameter vectors $\btheta_i,~i=1,\ldots,n$, can be approximated by the minimal Mahalanobis distance over the entire set of $\zeta$ views. The accuracy of this approximation depends on the number of available views. Algorithm~\ref{alg:MVICA} summarizes the intrinsic similarity of the approximation procedure in Proposition~\ref{lem:mhana_eucl}.
\begin{algorithm}[!h]
    \caption{Multi-view Intrinsic Similarity Approximation}
    \label{alg:MVICA}
    \KwIn{Data points: $\bx_1,...,\bx_n \in \mathbbm{R}^m$ divided into $\zeta$ views, $\varepsilon$}
    \KwOut{The approximated intrinsic kernel $\hat{K_\varepsilon}({\btheta_i},{\btheta_j}) $ and the approximated intrinsic similarity $d_{\btheta_i,\btheta_j}$.}
    \begin{algorithmic}[1]
    	  \STATE \For{$l=1$ to $\zeta$}{
    	 \hspace{0.2cm}	\For{ $i,j=1$ to $n$ }{
         \hspace{0.3cm} Compute  $C_{\btx_{i,l}}  \quad 1 \leq i \leq n $  using Eq.~\ref{eq:ito_corr} \\
         \hspace{0.3cm} Compute $d_l \left(\btx_{i,l},\btx_{j,l} \right) = \frac{1}{2} \left(\btx_{i,l} -
         \btx_{j,l} \right)^T  \left( C_{\btx_{i,l}}^{-1} + C_{\btx_{j,l}}^{-1} \right) \left(  \btx_{i,l} -  \btx_{j,l} \right)$ \\}}
         \STATE Compute $d_{\btheta_i,\btheta_j} = G \left( d_l \left(\btx_{i,l},\btx_{j,l} \right) \right)$ where $G\left( \right)$ is the minimization over $\zeta$ views \\
        \STATE $\hat{K_\varepsilon}({\btheta_i},{\btheta_j}) = exp\{-d_{\btheta_i,\btheta_j}/\varepsilon\}$
    \end{algorithmic}
\end{algorithm}

Algorithm~\ref{alg:MVICA} outputs  $\hat{K}_\varepsilon({\btheta_i},{\btheta_j}) \approx exp \{-\| {\btheta_i}-{\btheta_j} \|^2/\varepsilon\} $ if $\zeta$ is sufficiently large and its corresponding similarity $d_{\btheta_i,\btheta_j}$ can be utilized to approximate any kernel of the form of Eq.~\ref{eq:kernel_def} over the set of intrinsic parameters. It is important to note that for the computation of $d_l \left(\btx_{i,l},\btx_{j,l} \right)$, care should be taken to consider  only $ C_{\btx_{i,l}}$ and  $C_{\btx_{j,l}}$ that have at least rank of $d \leq k_l$. The  computational complexity of the $l$th covariance computation is $O \left(m_l^2 N  + m_l^3 \right)$, where $N$ is the number of neighbors per data point.
The computational complexity, which is required for the computation of  the $\zeta$ kernel matrices, is $O \left(\sum_{l=1}^\zeta m_l^2 n^2 \right)$ and in the worst case where $m_l \approx m$ we have   $O \left(\zeta m^2 n^2 \right)$.  Algorithm~\ref{alg:MVICA} may serve as a preprocessing step to the ICA procedure in \cite{singer:spectralICA} to reduce the contribution from the undesired interference.

%%%%%%%%%%%%%%%%%%%%%%%%%%%%%%%%%%%%%%%%%%%%%%%%%%%%%%%%%%%%%%%%%%
\section{Data with an accessible covariance matrix}
\label{sec:data_with_acc_cov}
%%%%%%%%%%%%%%%%%%%%%%%%%%%%%%%%%%%%%%%%%%%%%%%%%%%%%%%%%%%%%%%%%%
In this section, we propose to generate for  data $M$ a multi-view kernel $K_{\varepsilon}$ and a function $G$ from Eq.~\ref{eq:multi-view_kernel}. In this section, we assume that the covariance matrix at each data point in $M$ is accessible or can be computed using local neighborhood of each data point. Our goal is to approximate the kernel affinities between data points based on multiple views of the transformed
 intrinsic space. Following Section~\ref{sec:problem_Formulation}, we assume that the $l$th view is the outcome of an almost anywhere differentiable function  $\bbf_l:\Rn{d} \times \Rn{k_l} \to \Rn{m_l}$  such that $\btx_{i,l} = \bbf_l \left(\btheta_i, \bpsi_{i,l} \right)$. Everywhere in this section, we assume that $\btheta_i$ is not necessarily an It\^{o} process and  $k_l= 0$. Hence, a data point of each view   is strictly a function of the  intrinsic parameters in the consensus $\bbf_l:\Rn{d} \to \Rn{m_l}$ and  $\btx_{i,l}=\bbf_l \left(\btheta_i \right)$.

%We start with some background, denote the average value of the $i$-th point within the $l$-th view as
%$\eta_{i,l} \defeq (1/{m_l}) \sum_{b=1}^{m_l} \bx_i[b]$. We center the data using the following $\tilde{{x}}_{i,l} \defeq {x}_{i,l} - \eta_{i,l} \ldots {1}$ (where ${1}$ denotes a $m_l \times 1$ all-ones vector). The sample covariance matrix elements are computed using the following

The dimension $m_l$ of the $l$-view is assumed to be  higher than
the dimension $d$ of the intrinsic parametric space. Furthermore, the covariance matrix
maximal rank is equal to the intrinsic dimension $d$. Therefore, we use the
Moore-Penrose pseudoinverse to compute the Mahalanobis distance for $i,j=1, \ldots, n$,
$l=1, \ldots , \zeta$, by
\begin{equation}
\label{eq:mahal2}
d_m \left( \btx_{i,l}, \btx_{j,l} \right) = \frac{1}{2} \left( \btx_{i,l}  - \btx_{j,l} \right)^T \left(  C_{\btx_{i,l}}^{\dagger} +  C_{\btx_{j,l}}^{\dagger}  \right)  \left( \btx_{i,l}  - \btx_{j,l} \right).
\end{equation}
The Mahalanobis distance enables us to compare data points in the intrinsic space by comparing data points in the ambient space as Proposition~\ref{cor:mahal} suggests.
\begin{proposition}
    \label{cor:mahal}
    Let   $\bbf_l$ be a  bi-Lipschitz function such that $\btx_{i,l}=\bbf_l \left(\btheta_i \right)$ and $\btx_{j,l}=\bbf_l \left(\btheta_j \right)$ are two data points from $M_l$. Then, for $\phi \defeq \|\btheta_{i}-\btheta_{j}\|$
    \begin{equation}
    \label{eq:mahalDapp}
    d_m \left( \btx_{i,l}, \btx_{j,l} \right) = d_m \left( \btheta_{i}, \btheta_{j} \right)+O(\phi^4).
    \end{equation}
 \end{proposition}
    \begin{proof}
        The function $\bbf_l$ is expanded into first order Taylor near the point $\btheta_i$ such that\\
        \begin{equation}
        \label{eq:taylor}
        \btx_{i,l}=\btx_{j,l} + J_{\btx_{i,l}}(\btheta_j-\btheta_i) +O(\phi^2)
        \end{equation}
        where $J_{\btx_{i,l}}$ is the Jacobian of $\bbf_l \left(\btheta_i \right)$. By using Eqs. \ref{eq:taylor}  and \ref{eq:mahal2}, we get
        \begin{equation}
        \begin{array}{lll}
        d_m \left(\bbf_l \left(\btheta_i\right),\bbf_l \left(\btheta_j\right) \right) & = &  \\
        &&\frac{1}{2}  \left(\btheta_i - \btheta_j \right)^T  J_{\btx_{i,l}}^T \left( J_{\btx_{i,l}} C_{\btheta_i}    J_{\btx_{i,l}}^T\right)^{\dagger}    J_{\btx_{i,l}} \left(\btheta_i - \btheta_j \right) + \\
        && \frac{1}{2}  \left(\btheta_i - \btheta_j \right)^T  J_{\btx_{j,l}}^T
        \left( J_{\btx_{j,l}} C_{\btheta_j}    J_{\btx_{j,l}}^T\right)^{\dagger}
         J_{\btx_{j,l}} \left(\btheta_i - \btheta_j \right)+O(\phi^4).
        \end{array}
        \label{eq:maha2}
        \end{equation}
        The term $O(\phi^3)$ was canceled due to the symmetrization in Eq.~\ref{eq:maha2}.
        Assume that the rank of  $ J_{\btx_{i,l}}$ is equal to the rank of $C_{\btheta_i}$
        where $C_{\btheta_i}$ is a full rank. By using $m \geq d$ we get
        \begin{eqnarray}
        \label{eq:mahal3} d_m \left(\bbf_l \left(\btheta_i\right),\bbf_l
        \left(\btheta_j\right) \right) = \frac{1}{2}  \left(\btheta_i -
        \btheta_j \right)^T   C_{\btheta_i}^{-1} \left(\btheta_i - \btheta_j
        \right) +
        \frac{1}{2}  \left(\btheta_i - \btheta_j \right)^T  C_{\btheta_j}^{-1}   \left(\btheta_i - \btheta_j \right) +O(\phi^4)
        \end{eqnarray}
        that can be rewritten as
        \begin{eqnarray}
        \label{eq:mahal4} d_m \left( \btx_{i,l}, \btx_{j,l} \right) =d_m \left(\bbf_l \left(\btheta_i\right),\bbf_l
        \left(\btheta_j\right) \right) = \frac{1}{2}  \left(\btheta_i -
        \btheta_j \right)^T  \left( C_{\btheta_i}^{-1} +C_{\btheta_j}^{-1}
        \right) \left(\btheta_i - \btheta_j \right) +O(\phi^4) .
        \end{eqnarray}       
    \end{proof}
According to Proposition~\ref{cor:mahal}, for a small distance $\phi$, the Mahalanobis distance $d_m \left( \btx_{i,l}, \btx_{j,l} \right)$ is approximately the same in each view $l$ such that $d_m \left( \btx_{i,l_1}, \btx_{j,l_1} \right) \approx d_m \left( \btx_{i,l_2}, \btx_{j,l_2} \right),~l_1,l_2 = 1,\ldots,\zeta$.

%Note that for a given view $l$, if the $rank(J_{l}(\bx_i))$ is not equal to the $rank(C_{\theta})=d$, the above approximation does not hold. Based on this observation we claim that not only the intrinsic dimension could be approximated using the Mahalanobis distance \cite{Singer}, but the multiple views could be compared to find a better approximation of the affinities in the inaccessible parametric space.

The result proven in Proposition \ref{cor:mahal} is valid only if the function $\bbf_l$ is bi-Lipschitz and if the rank of  $ J_{\btx_{i,l}}$,  equals to the intrinsic dimension $d$. If the first assumption does not hold at point $\btx_{i,l}$, then the function $f_l$ is not bi-Lipschitz and thus for every $C>0$ there exists $\btheta_j$ such that $|f_l \left(\btheta_i \right)-f_l \left(\btheta_j \right)|>C|\btheta_i -\btheta_j |$. Hence, by the definition of $d_m$ in Eq.~\ref{eq:mahal2}, $   d_m \left( \btx_{i,l}, \btx_{j,l} \right)> d_m \left( \btheta_{i}, \btheta_{j} \right)  $.\\
If the second assumption does not hold at point $\btx_{i,l}$ then the rank of $ J_{\btx_{i,l}}$ is strictly smaller than $d$, therefore, if $ \| (\btheta_j-\btheta_i) \|\neq 0$ is in the null space of $J_{\btx_{i,l}}$, then $ \| J_{\btx_{i,l}}(\btheta_j-\btheta_i) \|=0$. Hence, $ d_m \left( \btx_{i,l}, \btx_{j,l} \right)=0$ although $d_m \left( \btheta_{i}, \btheta_{j} \right) \neq 0$. In the following, we suggest to validate the above two assumptions by finding the minimal distance while choosing the estimated covariances $ C_{\btx_{i,l}}$,  $1 \leq i \leq n$, $1 \leq l \leq \zeta$, that are used for distance estimation to have a local rank that is larger than the intrinsic dimension.

Estimating the intrinsic dimensionality $d$ of a data has recently gained considerable attention. Methods such as \cite{fukunaga1971algorithm,verveer1995evaluation} use local or global PCA to estimate the intrinsic dimension ${d}$. The dimension is set as the number of eigenvalues that are larger than some threshold. Others, such as \cite{trunk1976stastical,pettis1979intrinsic}, use K-NN distances to find a subspace around each data point and are based on some statistical assumption that estimates ${d}$. A survey of different approaches is presented in \cite{camastra2003data}. The intrinsic dimensionality is estimated from Angle and Norm Concentration (DANCo) in~\cite{DANCo}. In the following, we utilize the local PCA to estimate $d$~\cite{singer:ODM}. This method is  integrated into  Algorithm~\ref{alg:Mahalan}.

Algorithm~\ref{alg:Mahalan} approximates the intrinsic dimension $d$ and provides an improved affinity measure that can be used for various kernel-based methods such as DM that reveals the underlined manifold.
This algorithm identifies deviation from the assumptions that  $\bbf_l$ is bi-Lipschitz and if the rank of  $ J_{\btx_{i,l}}$,  equals to the intrinsic dimension $d$.

\begin{algorithm}[h!]
    \caption{Multi-view affinity measure approximation}
    \label{alg:Mahalan}
    \KwIn{Data points: $\bx_1,...,\bx_n \in \mathbbm{R}^m$ divided into $\zeta$ views, associated covariances $C_{\btx_{i,l}}$ $1 \leq i \leq n$,  $1 \leq l \leq \zeta$, $\varepsilon$, threshold $\gamma$}
    \KwOut{Approximated $K_\varepsilon({\btheta_i},{\btheta_j})$}
    \begin{algorithmic}[1]
        \STATE Compute $\kappa \left(i,l \right)$ as the number of singular values of $C_{\btx_{i,l}}$ that are larger than $\gamma$
        \STATE Compute $\kappa_m$ as the median of $\kappa \left(i,l\right)$, $1 \leq i \leq n$, $ 1 \leq l \leq  \zeta$
        \FOR{ $ i,j= 1$ to $n$, $l=1$ to $\zeta$, where $\kappa(i,l) \geq \kappa_m$ and  $\kappa(j,l) \geq \kappa_m$}
        \STATE Calculate $d_m \left( \btx_{i,l}, \btx_{j,l} \right)$ using Eq.~\ref{eq:mahal2}
        \STATE Set $K_{\varepsilon} ({ \btx_{i,l}},{ \btx_{j,l}})=exp \left( {-d_m \left( \btx_{i,l}, \btx_{j,l} \right)/\varepsilon} \right)$ 
        \STATE Set $K_{\varepsilon}({\btheta_i},{\btheta_j}) =  G(K_{\varepsilon} ({ \btx_{i,1}},{ \btx_{j,1}}),\ldots,K_{\varepsilon} ({ \btx_{i,\zeta}},{ \btx_{j,\zeta}}))$ where $G\left( \right)$ is the  majority vote operation
        \ENDFOR
    \end{algorithmic}
\end{algorithm}%%%%%%%%%%%%%%%%%%%%%%%%%%%%%%%%%%% It\^{o}

The first step in Algorithm~\ref{alg:Mahalan}  computes the local covariance with computational complexity of $O \left( N m_l^2 \right)$, where $N$ is the number of neighbors and $m_l$ is the dimension of $C_{\btx_{i,l}}$. Step $1$ is applied $n$ times. In the second step, the algorithm computes the numerical rank of  $C_{\btx_{i,l}}$, $1 \leq i \leq n$. This step has a computational complexity cost of $O \left( n m_l^3\right)$. In step $3$, the intrinsic dimension is estimated by the median of $\kappa(j,l)$, $1 \leq i \leq n$, $ 1 \leq l \leq  \zeta$, with computational complexity of $O \left( n \zeta \right)$. In step $4$ computes the kernel. This is achieved by computing first the Mahalanobis distance  $d_m \left( \btx_{i,l}, \btx_{j,l} \right)$, $1 \leq i \leq n$, $ 1 \leq l \leq  \zeta$,  with a total computational complexity $O \left( n \sum_{l=1}^{\zeta} m_l^3\right)$. The distances are computed between data points that correspond to covariances with sufficiently large numerical rank. The output from the $i,j$th element in the kernel is estimated as the output from the majority vote using the function $G\left( \right)$ over the kernel elements distances for the relevant views.
In cases where  the bi-Lipschitz condition is violated for a small number of data points, the function $G \left( \right)$ can include an histogram analysis to find the maximal accumulation point. According to Proposition~\ref{cor:mahal}, this maximal accumulation point is related to the true intrinsic distance up to a small error.  

Algorithm~\ref{alg:Mahalan} is similar to Algorithm~\ref{alg:MVICA} in the sense that both are based on computing the Mahalanobis distance for relevant covariances. However, the role of the function $G \left( \right)$ is very different. The function $G \left( \right)$ in Algorithm~\ref{alg:MVICA}, aims to find the less-noisy distance as the minimal distance over the views per data point. On the other hand, the function $G \left( \right)$ in Algorithm~\ref{alg:Mahalan},  aims to find the most agreeable distance over the views by using a majority vote function. Another interesting difference lies in the computation of the covariances. Algorithm~\ref{alg:MVICA}, assumes that the samples are from a stochastic process and can be used to compute the covariances.  Algorithm~\ref{alg:Mahalan} assumes that the covariances are given as inputs.
%%%%%%%%%%%%%%%%%%%%%%%%%%%%%%%%%%%%%%%%%%%%%%%%%%%%%%%
\section{Experimental Results}
\label{sec:examples}
%%%%%%%%%%%%%%%%%%%%%%%%%%%%%%%%%%%%%%%%%%%%%%%%%%%%%%%
This section describes three examples that demonstrate how the multi-view approaches are used. The first example (Section~\ref{sec:examples_full}) describes an embedding of data that consists of several It\^{o} processes in the presence of noise. We show that we can single out the consensus from noisy measurements using the method in Section~\ref{sec:Multi-View of Dynamical Process}.  The second example (Section~\ref{sec:MVCA}) describes data embedding that consists of several views with an accessible covariance matrix. In this case, we show the advantage of the multi-view-based embedding from Section~\ref{sec:data_with_acc_cov}. It is compared to the embedding of a corresponding single-view dataset where its features are concatenated of all the features subsets from the views. The third example (Section~\ref{sec:classification_of_sesmic_events}) compares the method from  Section~\ref{sec:data_with_acc_cov} with the state-of-the-art method~\cite{MV1} for a classification of real-life seismic events data.

%%%%%%%%%%%%%%%%%%%%%%%%%%%%%%%%%%%%%%%%%%%%%%%%%%%%%%%%
\subsection{Noisy Multi-Views with Consensus: Example I}
\label{sec:examples_full}
%%%%%%%%%%%%%%%%%%%%%%%%%%%%%%%%%%%%%%%%%%%%%%%%%%%%%%%%

\begin{figure}[H]
    \centering
    %\vspace*{1.5cm}\hspace*{0.5cm}\rotatebox{90}{\textcolor{black}{$Q$}}

    \includegraphics[width=0.9\textwidth] {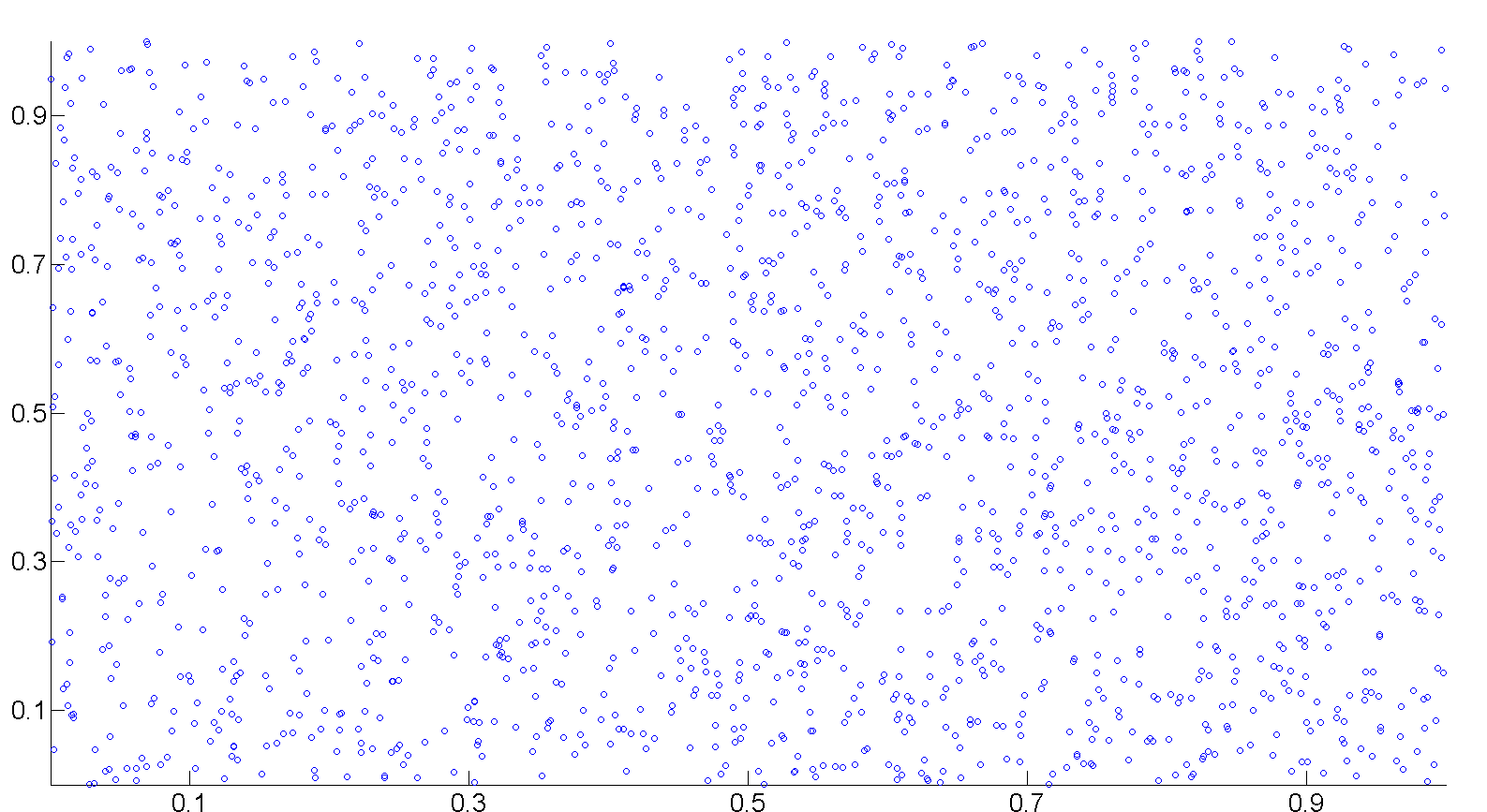}
    \vspace*{1.5cm}\hspace*{8.5cm}\textcolor{black}{$\theta_i^2$}
     \leavevmode\smash{\makebox[0pt]{\hspace{-42em}% HORIZONTAL POSITION
    \rotatebox[origin=l]{90}{\hspace{10em}% VERTICAL POSITION
    $\theta_i^1$}%
}}\hspace{0pt plus 1filll}\null
    \caption{The intrinsic parameters $\btheta_i,~i=1,\ldots,2000$, of the data.}
    \label{fig:consensus}
\end{figure}

For the numerical example of the process described in Section~\ref{sec:Multi-View of Dynamical Process}, we consider the Brownian motion $(\theta^1_i, \theta^2_i) = (w^1_i, w^2_i),~i=1,\ldots,2000$,  in the unit square $[0, 1]\times[0, 1]$ with a normal reflection at the boundary. Furthermore, the interferences $\bpsi_{i,l}~i=1,\ldots,2000$, are Brownian motion in the interval  $[0, 1]$ with a normal reflection at the boundary.  The $n=2000$ data points are sampled uniformly from the unit square as depicted in Fig.~\ref{fig:consensus}.  The intrinsic parameters $\btheta_{i}$ are measured via  the set of $\zeta$ views where $\zeta = 1, \ldots, 10$. Each view is a function of the consensus $\btheta_{i}$ and a function of an additional view-specific  It\^{o} process  given by  $\bpsi_{i,l}$, $1 \leq l \leq \zeta$, that are considered as interferences. 
%
%In the following, we consider the Brownian motion $(\theta^1_i, \theta^2_i) = (w^1_i, w^2_i),~i=1,\ldots,n$  in the unit square $[0, 1]\times[0, 1]$ with a normal reflection at the boundary. We observe  $(\theta^1_i, \theta^2_i)$ through a non-linear mapping in the presence of an additional It\^{o} process that is not common to all of the views and considered as noise.
%%\begin{equation}
%\bx_i = \sum_{j=1}^3  a_{j,l} \btheta_j ^{b_j,l}, \quad i=1,..,3
%\end{equation}

%\begin{figure}[H]
%    \centering
% \includegraphics[width=0.45\textwidth] {consensus.png}
%    \includegraphics[width=0.45\textwidth] {brown_example.png}
%    \caption{Intrinsic parameters of the data and Transformed data of a specific example view.}
%    \label{fig:helix1}
%\end{figure}

%\begin{figure}[H]
%\centering
%\subfigure[ The intrinsic parameters $\btheta_i$ of the data.]{\label{fig:consensus} \includegraphics[width=0.47\textwidth] {consensus_c.png}}
%\subfigure[ A transformed $\btheta_i$ from a specific view $\btx_{i,l}$.] {\label{fig:brown_example} \includegraphics[width=0.47\textwidth] {brown_example_c.png}}
%\caption{Example of the single-view transformation of the Intrinsic parameters of the data.}
%\label{fig:given_data}
%\end{figure}

\begin{figure}[H]
    \centering
    %\vspace*{1.5cm}\hspace*{0.5cm}\rotatebox{90}{\textcolor{black}{$Q$}}

   \includegraphics[width=0.9\textwidth] {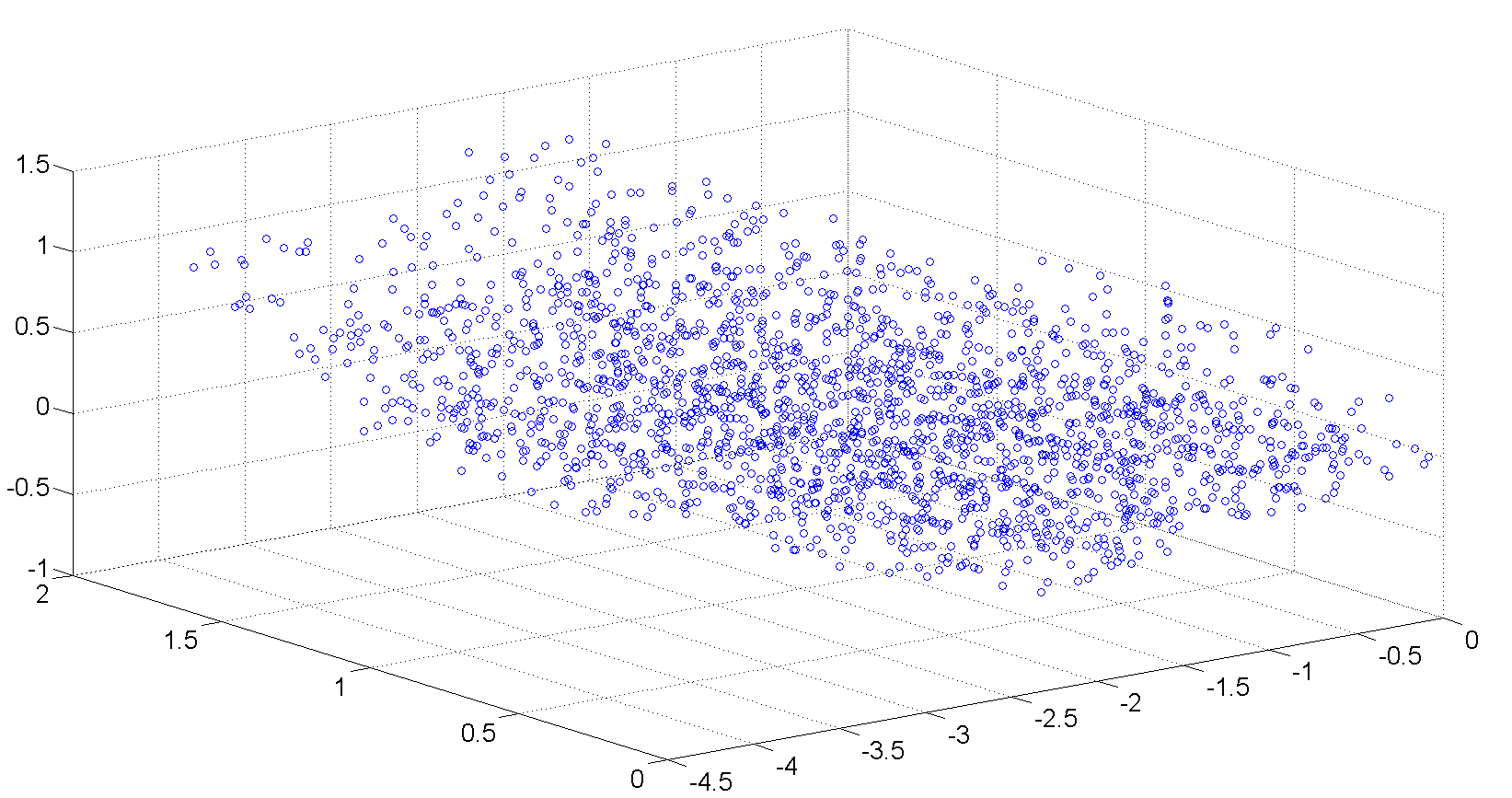}
    \vspace*{1.5cm}\hspace*{4.5cm}\textcolor{black}{$\tx_{i,l}^1~~~~~~~~~~~~~~~~~~~~~~~~~~~~~~~~$  $\tx_{i,l}^2$}
     \leavevmode\smash{\makebox[0pt]{\hspace{-50em}% HORIZONTAL POSITION
    \rotatebox[origin=l]{90}{\hspace{12em}% VERTICAL POSITION
    $\tx_{i,l}^3$}%
}}\hspace{0pt plus 1filll}\null
    \caption{ A transformed $\btheta_i$ from a specific view $\btx_{i,l}$ according to Eq.~\ref{eq:trans_points}}
    \label{fig:brown_example}
\end{figure}

%\begin{figure}[ht]
%    \centering
%    \includegraphics[width=0.9\textwidth] {consensus.png}
%    \caption{Intrinsic parameters of the data.}
%    \label{fig:consensus}
%\end{figure}
%
%\begin{figure}[h]
%    \centering
%    \includegraphics[width=0.9\textwidth] {brown_example.png}
%    \caption{Transformed data of a specific example view. }
%    \label{fig:brown_example}
%\end{figure}

%\theta_{i}

\begin{figure}[H]
    \centering
    %\vspace*{1.5cm}\hspace*{0.5cm}\rotatebox{90}{\textcolor{black}{$Q$}}

   \includegraphics[width=0.9\textwidth] {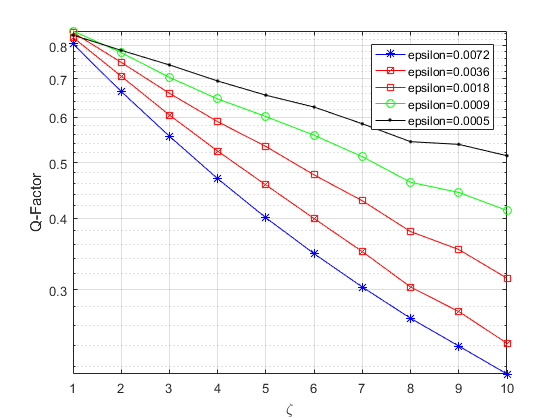}
       \leavevmode\smash{\makebox[0pt]{\hspace{-40em}% HORIZONTAL POSITION
}}\hspace{0pt plus 1filll}\null
    \caption{Q-factor as a function of $\zeta$. For each value of $\zeta$, the average accuracy is computed over $20$ simulation repetitions for several $\varepsilon$ values.}
    \label{fig:mv_accuracy}
\end{figure}

 We  measure each view $\btx_{i,l},~i=1,\ldots,2000$, through a non-linear transformation by
\begin{align}
\label{eq:trans_points}
\tx^{1}_{i,l} =     a_{1,1,l}(\theta_i^{1})^{b_{1,1,l}} + a_{1,2,l}(\theta_i^{2})^{b_{1,2,l}} + a_{1,3,l}(\psi_{i,l})^{b_{1,3,l}} \\ \nonumber
\tx^{2}_{i,l} =     a_{2,1,l}(\theta_i^{1})^{b_{2,1,l}} + a_{2,2,l}(\theta_i^{2})^{b_{2,2,l}} + a_{2,3,l}(\psi_{i,l})^{b_{2,3,l}} \\ \nonumber
\tx^{3}_{i,l} =     a_{3,1,l}(\theta_i^{1})^{b_{3,1,l}} + a_{3,2,l}(\theta_i^{2})^{b_{3,2,l}} + a_{3,3,l}(\psi_{i,l})^{b_{3,3,l}},
\end{align}
where $l=1,\ldots,\zeta$ and $a_{k,q,l}\neq0$ is random number chosen uniformly from the interval $[-2, 2]$, $k,q=1,\ldots,3$. Furthermore, $b_{k,q,l}$ was fixed to $3$. We call $\btx_{i,l}= (\tx^{1}_{i,l},\tx^{2}_{i,l},\tx^{3}_{i,l})$, $1 \leq i \leq 2000$, the observable sample from the space of the $l$th view. This sample is a function of the unknown controlling parameters $\theta^1_i$ and $\theta^2_i$  and the unknown interference $\psi_{i,l}$ where $\btx_{i,l} = \bbf_l (\theta^1_i,\theta^2_i,\psi_{i,l})$.  An illustration of $\btx_{i,l}$ for a specific view is given in Fig.~\ref{fig:brown_example}. The specific instances of the parameters of the transformation (Eq.~\ref{eq:trans_points}), which are used in Fig.~\ref{fig:brown_example}, are
\begin{align}
    \tx^{1}_{i,l} &=  -1.94(\theta_i^{1})^3    + 0.24(\theta_i^{2})^{3} - 0.62(\psi_{i,l})^3 \\ \nonumber
    \tx^{2}_{i,l} &=  -1.59(\theta_i^{1})^{3}  + 1.39(\theta_i^{2})^{3}  + 0.53(\psi_{i,l})^{3} \\ \nonumber
    \tx^{3}_{i,l} &=  -0.68(\theta_i^{1})^{3} + 0.34(\theta_i^{2})^{3}  + 1.10(\psi_{i,l})^{3}.
\end{align}
Hence, the input data to Algorithm~\ref{alg:MVICA} is the concatenation of  $(\tx^{1}_{i,l},\tx^{2}_{i,l},\tx^{3}_{i,l})$  $\zeta$ times such that $\bx_i = \cup_{l=1}^\zeta \btx_{i,l}$.  Algorithm~\ref{alg:MVICA} was applied to estimate the pairwise intrinsic distances between data points of the consensus. Initially, we run $N_c=20000$ stochastic simulations for a short time period $dt=0.005$  that were initiated at point $\btheta_i$. The result of the $N_c$ stochastic simulations is a point-cloud in the neighborhood of $\btheta_i$. Each point-cloud is mapped using the transformation in Eq.~\ref{eq:trans_points} to a point-cloud in the neighborhood of $\btx_{i,l}$ in the measured space. Following Eq.~\ref{eq:ito_corr_jac}, we calculate the $3\times 3$ sample  covariance matrix for the point-cloud of each data point in each view. Then, the Mahalanobis distance from Eq.~\ref{eq:ica_dist} is used to estimate the  pairwise distances between intrinsic data points in the consensus plus an interference contributed factor. The minimal Mahalanobis distance (over the views) is used to estimate the  pairwise distance between intrinsic data points of the consensus $\btheta_i$ and $\btheta_j$ $i,j=1,\ldots,n$, by
\begin{equation}
\label{eq:min_zeta_kernel}
\hat{K}_\varepsilon(\btheta_i ,\btheta_j ) \approx \min_{1\leq l \leq \zeta} \left\{\exp{\frac{-||\btheta_i - \btheta_j||^2-||\bpsi_{i,l} - \bpsi_{j,l}||^2}{\varepsilon}}\right\},
\end{equation}
for several $\varepsilon$ values. According to Proposition~\ref{lem:mhana_eucl}, when more views are given, the accuracy of the estimated pair-wise affinities improves. Let $K_\varepsilon(\btheta_i ,\btheta_j) = \exp{\frac{-||\btheta_i-\btheta_j||^2}{\varepsilon}}$ be the inaccessible and anisotropic ground truth kernel. The quality factor $Q$ of the $K_\varepsilon$ approximation  of $\hat{K}_\varepsilon$ is measured by the relative error in Frobenius norm
\begin{equation}
Q = \frac{||K_\varepsilon-\hat{K}_\varepsilon||_F}{||\hat{K}_\varepsilon||_F}.
\end{equation}
Figure~\ref{fig:mv_accuracy} displays the relation between the number of views and the approximation quality  factor $Q$ of the kernel estimation for several $\varepsilon$ values. The $Q$-Factor becomes smaller with increasing $\varepsilon$ since $||K_\varepsilon||_F$ grows faster than $||K_\varepsilon-\hat{K}_\varepsilon||_F$. This is the result of a reduced noise contribution  $||\bpsi_{i,l} - \bpsi_{j,l}||^2/\varepsilon$ in Eq.~\ref{eq:min_zeta_kernel} as $\varepsilon$ grows while $||K_\varepsilon||_F$ and $||\hat{K}_\varepsilon||_F$ grow larger.

\begin{figure}[H]
	\centering
	\includegraphics[width=0.9\textwidth] {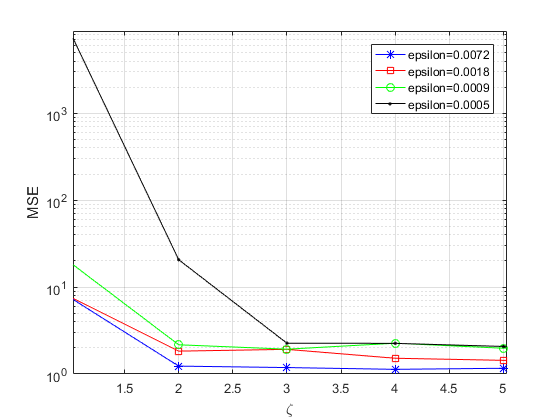}
	\caption{MSE of the Foker-Plank eigenvalues on the unit square in the presence of noisy multi-view process as a function of $\varepsilon$.}
	\label{fig:mse_spectral_lines}
\end{figure}

\begin{figure}[H]
    \centering
    \includegraphics[width=1.1\textwidth] {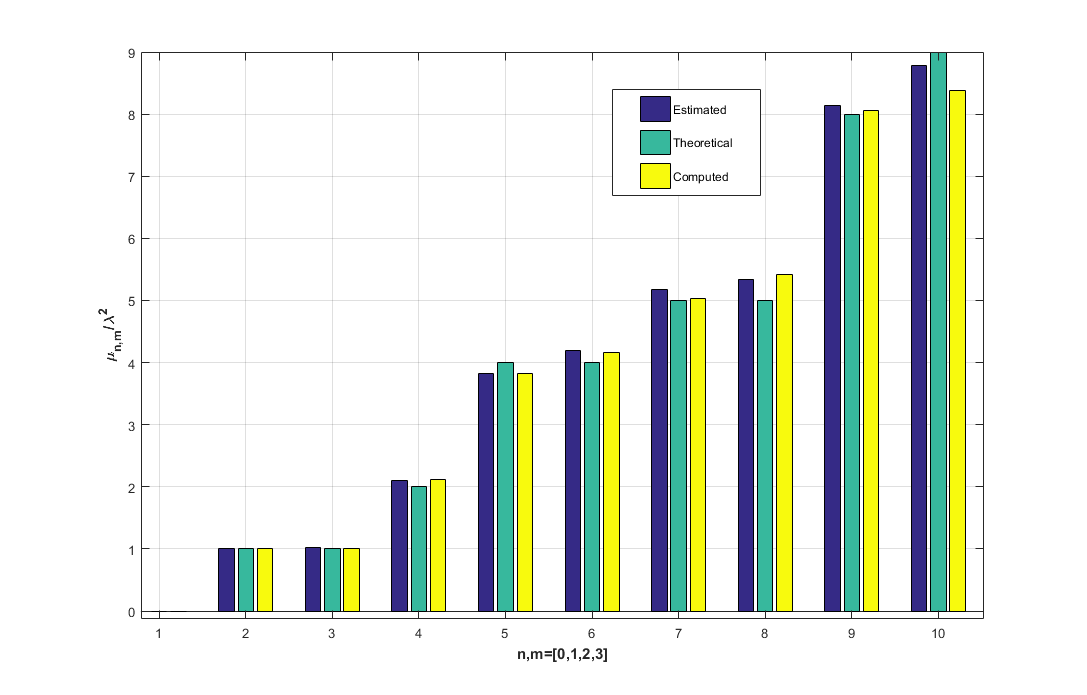}
    \caption{Recovery of the Foker-Plank eigenvalues on the unit square in the presence of noisy multi-view process: theoretical spectral lines (green), the approximated spectral lines (blue) and the computed spectral lines using $K_{\varepsilon}$ (yellow), $\varepsilon=0.0072$}
    \label{fig:spectral_lines}
\end{figure}

According to~\cite{singer:non-linearICA}, the anisotropic kernel $K_\varepsilon$ converges to Fokker-–Planck operator. Hence,  $ D^{-1}K_\varepsilon \approx -\frac{\varepsilon}{2} L$, where $D$ is a diagonal matrix  with $\left [ D \right]_{ii} = \sum_{j=1}^n {K}_\varepsilon(\btheta_i ,\btheta_j ) $ and $L$ is the Fokker–-Planck operator of the parametric manifold $\theta_i^1$ and $\theta_i^2$ in Fig.~\ref{fig:consensus}. From the definition of $\btheta_i$, in this  example, the parametric manifold is the unit square with a uniform density, so that $L$ is the Laplacian of the
unit square whose eigenvalues (for the Neumann boundary conditions) are $\mu_{n,m} = \pi^2 \left(n^2+m^2 \right),~n,m=0,1,\ldots$. Therefore, the eigenvalues $\lambda_i$ of the row stochastic matrix $D^{-1}K_\varepsilon$ are given by $\lambda_i \approx e^{-\frac{\varepsilon}{2} \mu_{n,m}}$. Figure~\ref{fig:spectral_lines} shows the values of $-2\log \left(\lambda_i \right)/\left(\pi^2 \varepsilon \right),~i=1,\ldots,10$, for the approximated $ \hat{D}^{-1}\hat{K}_\varepsilon$, where $\hat{D}$ is a diagonal matrix  and $\left [ \hat{D} \right]_{ii} = \sum_{j=1}^n\hat{K}_\varepsilon(\btheta_i ,\btheta_j ) $. The approximation accuracy of the spectral lines $n^2+m^2 = \left[0,1,1,2,4,4,5,5,8,9\right]$ are shown in Figs~\ref{fig:mse_spectral_lines} and~\ref{fig:spectral_lines}. Figure~\ref{fig:mse_spectral_lines} presents the Minimum Squared Error (MSE) between the theoretical spectral lines and the approximated ones as a function of $\zeta$,  where the approximated $\lambda_i~i=1,\ldots,10$, were computed using the multi-view scheme of Algorithm~\ref{alg:MVICA} for several $\varepsilon$ values averaged over $20$ randomization of the transformation from Eq.~\ref{eq:trans_points}. As seen in Fig.~\ref{fig:mse_spectral_lines}, the accuracy improves significantly when the number of views increases from $1$ to $2$ views and from $2$ views to $3$ views. Figure~\ref{fig:spectral_lines} compares the estimated  spectral lines $n^2+m^2$ using $10$ views with the theoretical ones and the computed ones using $K_{\varepsilon}$ directly. Figure~\ref{fig:spectral_lines} shows that although the parametric manifold was sampled in the presence of noise and an unknown non-linear transformation, the approximation is done with relatively small MSE of $0.55$ between the estimated spectral lines and the theoretical one when the first non-zero spectral line is normalized to $1$.

\subsection{Multi-views with an accessible covariance matrix: Example II}
\label{sec:MVCA}
\begin{figure}[H]
\centering
%\figure[ Helix I (Eq. \ref{eq: helixI})]{\label{fig:helixI}} 
%\subfigure[ Helix II (Eq. \ref{eq: helixII}).] {\label{fig:helix2} \includegraphics[width=0.4\textwidth] {Helix2.eps}}
{\label{fig:helixI}}
\includegraphics[width=0.7\textwidth]{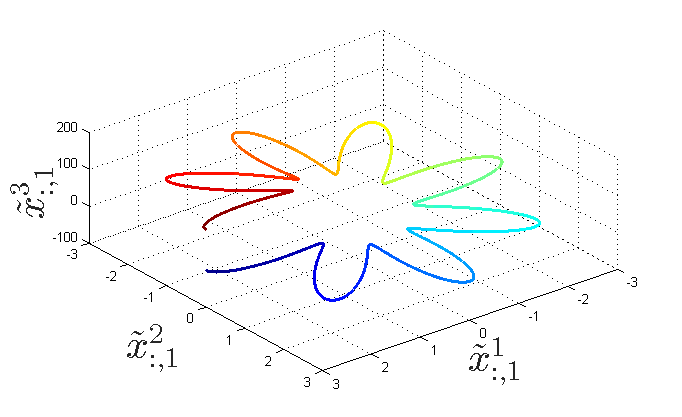}
    \caption{The sampled helix (Eq. \ref{eq: helixI})}
    \label{fig:helix1}
\end{figure}

Prior to performance demonstration of Algorithm~\ref{alg:Mahalan}, we evaluate the approximation of the Mahalanobis distance in the inaccessible feature space using Eq.~\ref{eq:mahalDapp}. First, we generate a $3$-dimensional manifold using the following equation
\begin{equation}
\label{eq: helixI}
\text{Helix :   }
\btx_{i,1}=
\begin{bmatrix}
{\tx^1_{i,1}}\\
{\tx^2_{i,1}}\\
{\tx^3_{i,1}}\\

\end{bmatrix}
=
\begin{bmatrix}
{f^1_1}(\theta_i)\\
{f^2_1}(\theta_i)\\
{f^3_1}(\theta_i)\\

\end{bmatrix}
=
\begin{bmatrix}
{(2+\cos(8\theta_i))\cdot \cos(\theta_i)}\\
{(2+\cos(8\theta_i))\cdot \sin(\theta_i)}\\
{(3\theta_i^2-\theta_i)}\\

\end{bmatrix},
\end{equation}
where the intrinsic governing parameters $\theta_i$, $i=1,\ldots,n=20,000$, are generated by $\theta_i$ that are uniformly drawn  from $[0,2\pi]$. The generated manifold is presented in Fig.~\ref{fig:helix1}.

%\begin{figure}[H]
%    \centering
% \includegraphics[width=0.4\textwidth] {Helix1.eps}
%  \includegraphics[width=0.4\textwidth] {Helix2.eps}
%    \caption{Helix I (Eq. \ref{eq: helixI}) and Helix II (Eq. \ref{eq: helixII}).}
%    \label{fig:helix1}
%\end{figure}

\begin{figure}[H]
\centering
%\subfigure[ Mahalanobis distance error for Helix I ]{\label{fig:MErrorH1} }
	\includegraphics[width=1.1\textwidth] {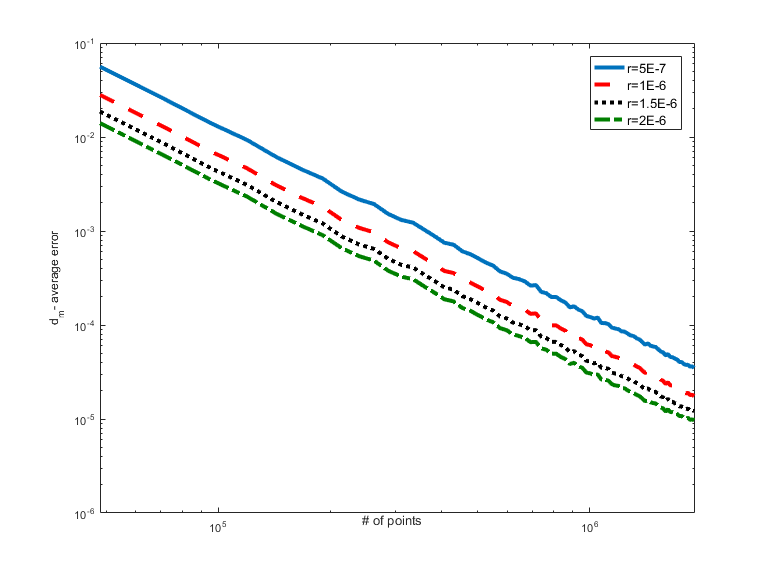}
	{\label{fig:MErrorH1} }
%\subfigure[ Mahalanobis distance error for Helix II] {\label{fig:MErrorH2} \includegraphics[width=0.6\textwidth] {MHErrorH2.eps}}
    \caption{The average  error between the Mahalanobis distances in the ambient space and the corresponding  Mahalanobis distances in the parametric space as a function of the number of points per unit length generated in the parametric space. The value $r$, represents the maximal distance that is used to find neighbors for the covariance matrix estimation.}
    \label{fig:ErrrorHelixRandom}
\end{figure}

%\begin{figure}[H]
%    \centering
%    \includegraphics[width=0.4\textwidth] {ErrorHelixI.eps}
%        \includegraphics[width=0.4\textwidth] {ErrorHelixII.eps}
%    \caption{The average  error between the Mahalanobis distance in the ambient space and between the equivalent in the parametric space. Simulated for a uniform grid and a random generated grid.}
%    \label{fig:ErrrorHelixRandom}
%\end{figure}
The empirical covariance matrix in Eq.~\ref{eq:mahal2} depends on the neighbors of each data point. Denote by $\varepsilon_r > 0$ the  neighborhood  radius used by the covariance matrix computation  at each data point.
The approximation average error in Eq.~\ref{eq:mahal2} is presented in Fig.~\ref{fig:ErrrorHelixRandom}, where the average is taken over $100$ repeated simulations. It is evident that the errors for both datasets diminish as  $\varepsilon_r $ grows and the number of neighbors increases. As a result,  the covariance matrix approximation improves and with it the similarity between both Mahalanobis distances in the ambient space. The corresponding Mahalanobis distances in the parametric space are improved.

\begin{figure}[H]
\centering
\subfigure[View I ]{\label{fig:Man1} \includegraphics[width=0.45\textwidth] {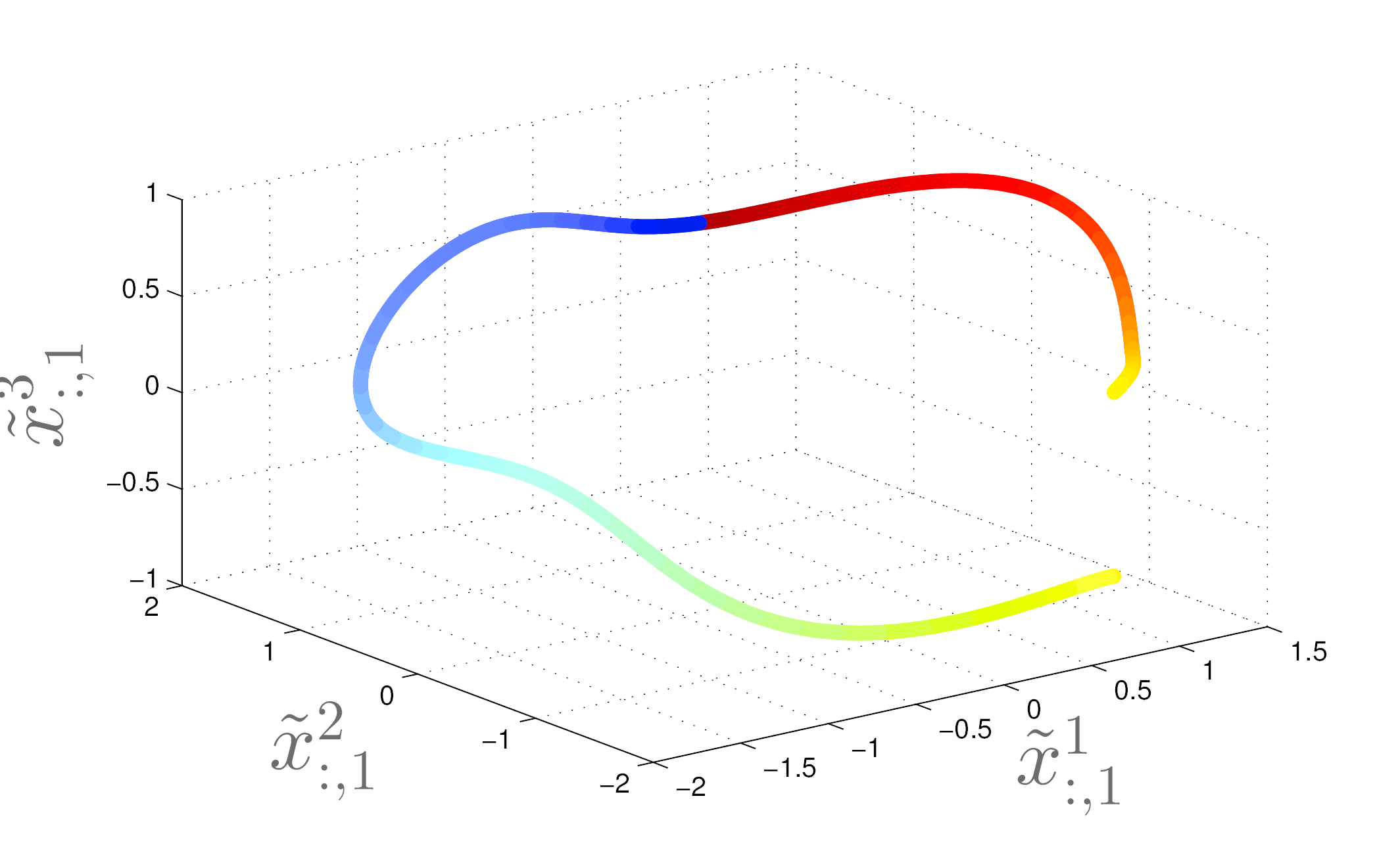}}
\subfigure[View II] {\label{fig:Man2} \includegraphics[width=0.45\textwidth] {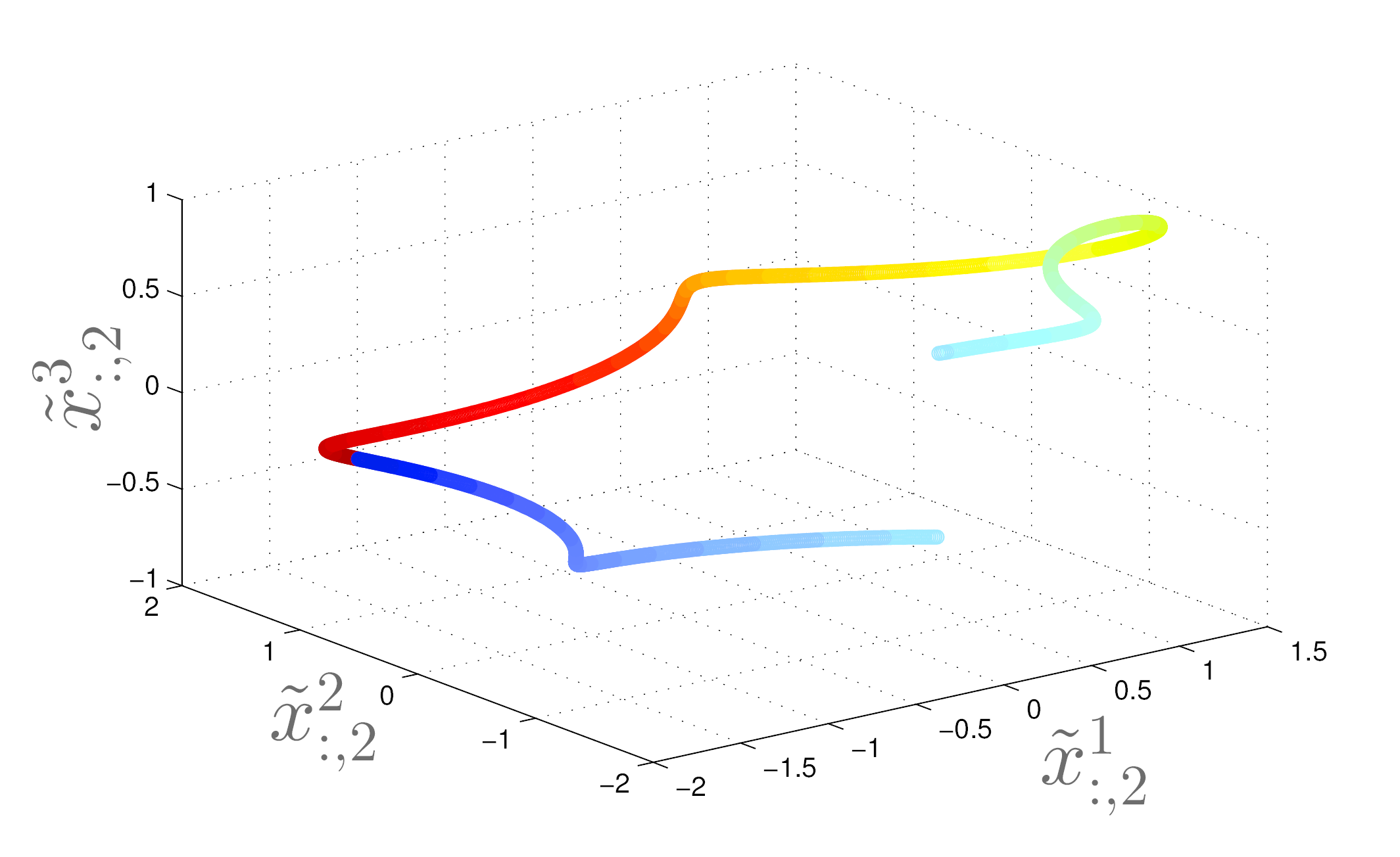}}
\subfigure[View III] {\label{fig:Man3} \includegraphics[width=0.45\textwidth] {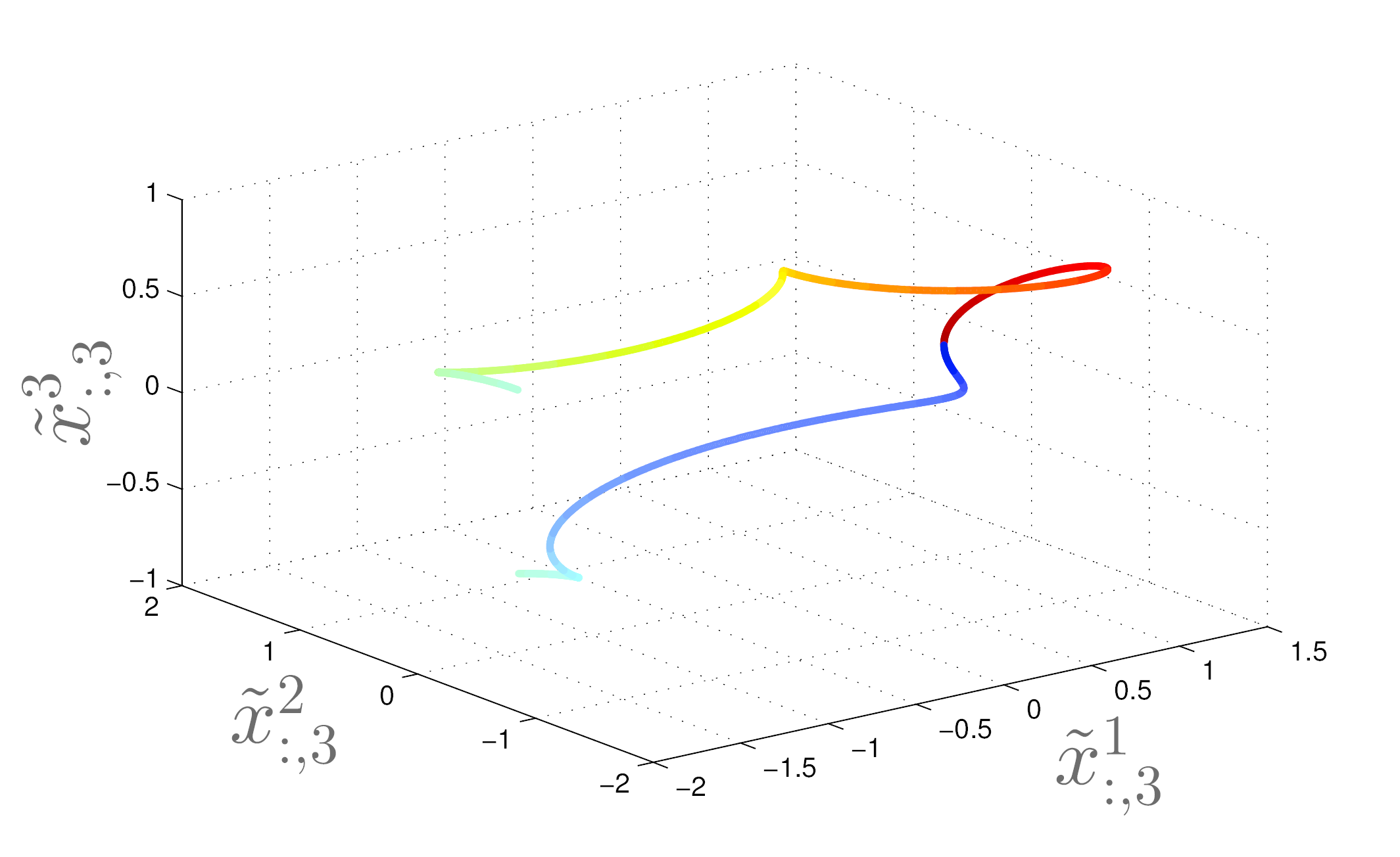}}
    \caption{The three manifolds generated by Eq.~\ref{eq: helixIII}.}
    \label{fig:helix3}
\end{figure}

In the rest of this section, the approximation of the affinity measure is evaluated.
$\zeta = 10$ views are generated to evaluate the performance of Algorithm~\ref{alg:Mahalan}. All the views are $3$-dimensional that are based on one underlying angular parameter denoted by $\theta_i \in \Rn{},~i=1,\ldots,n$. The $10$ views are generated by the application of the following function
\begin{equation}
\label{eq: helixIII}
\text{Helix III: }
\btx_{i,l}=
\begin{bmatrix}
{\tx_{i,l}^1}\\
{\tx_{i,l}^2}\\
{\tx_{i,l}^3}\\

\end{bmatrix}
=
\begin{bmatrix}
{f_l^1}(\btheta_i)\\
{f_l^2}(\btheta_i)\\
{f_l^3}(\btheta_i)\\

\end{bmatrix}
=
\begin{bmatrix}
{(4/3)\cdot \cos(\btheta_i+Z_l^1) -(1/3)\cdot \cos(4(\btheta_i+Z_l^1))}\\
{(4/3)\cdot \sin(\btheta_i+Z_l^2) -(1/3)\cdot \sin(4(\btheta_i+Z_l^2))}\\
{\sin(0.8\cdot\mod((\btheta_i+Z_l^3),2\pi))}\\

\end{bmatrix},
\end{equation}
where $Z_l^1,Z_l^2,Z_l^3, 1\leq l \leq 10$, are random variables
drawn from a uniform distribution on the interval
$[0,2\pi]$. As demonstrated in Fig.~\ref{fig:helix3}, each view is a
deformation of an open flower-shaped manifold. A similar deformation can occur
in various real-life applications where the measured data is the output from some non-linear
phenomena. In this experiment, we demonstrate the ability of a
multi-view approach to overcome such deformations.

\begin{figure}[H]
\centering
\subfigure[DM of View I ]{\label{fig:DMman1}  \includegraphics[width=0.45\textwidth] {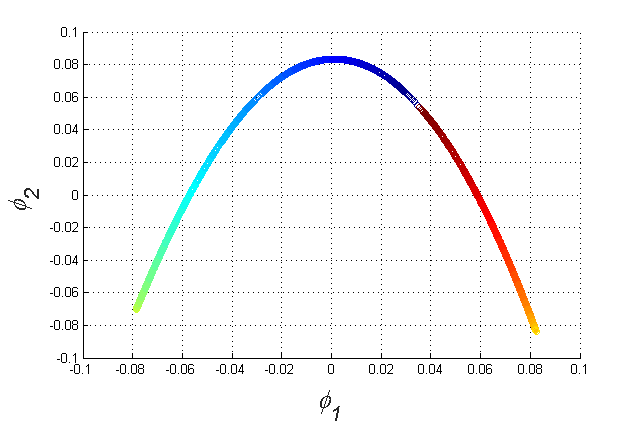}}
\subfigure[DM of View II] {\label{fig:DMman2} \includegraphics[width=0.45\textwidth] {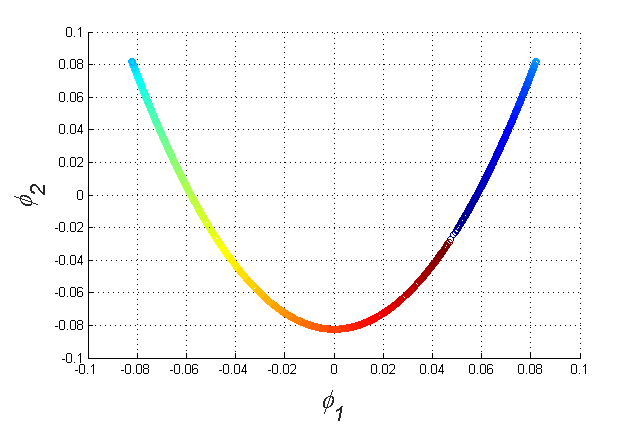}}
\subfigure[DM of View III] {\label{fig:DMman3} \includegraphics[width=0.45\textwidth] {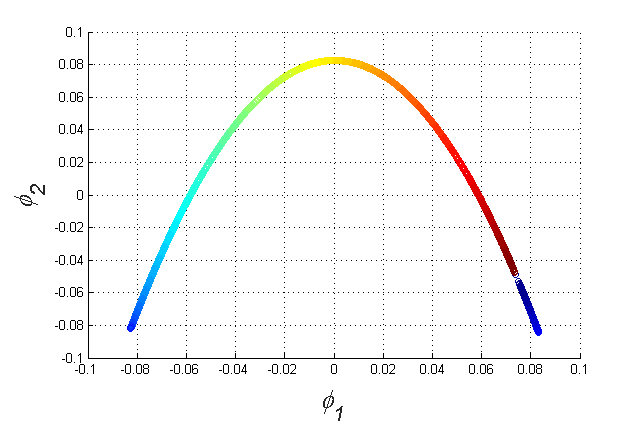}}
    \caption{The DM embedding of each single view $l=1, \dots, 3$.}
        \label{fig:helixSVDM}
\end{figure}

%\begin{figure}[H]
%    \centering
%    \includegraphics[width=0.45\textwidth] {Man1.eps}
%    \includegraphics[width=0.45\textwidth] {Man2.eps}
%        \includegraphics[width=0.45\textwidth] {Man3.eps}
%    \caption{Three manifolds generated using (Eq. \ref{eq: helixIII}) .}
%    \label{fig:helix3}
%\end{figure}
To extract the underlying parameter $\theta_i$, we first apply the  DM  to each view. The two leading coordinates of the extracted embedding are denoted by $\phi_1$ and $\phi_2$. They are presented in Fig.~\ref{fig:helixSVDM}. All the extracted manifolds are horseshoe shapes that consist of a large gap created by the deformation (Eq.~\ref{eq: helixIII}) in the third coordinate of the sampled data.

\begin{figure}[H]
\centering
\subfigure[DM cooredinates of the concatenated views]{\label{fig:DMconc} \includegraphics[width=0.32\textwidth] {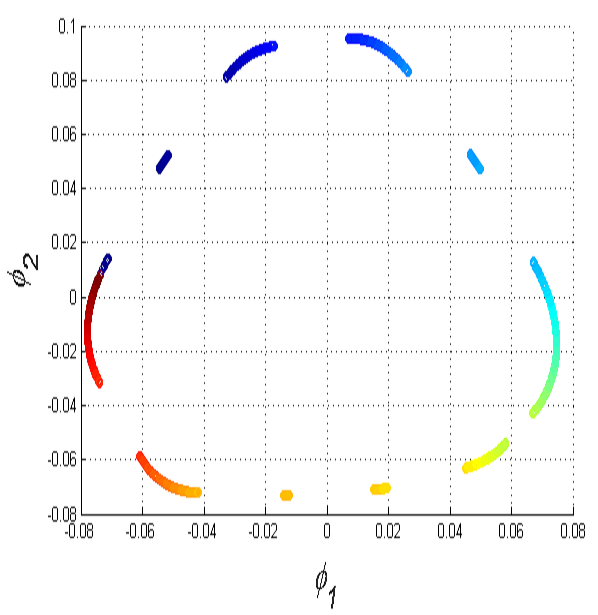}}
\subfigure[DM coordinates of  the multi-view-based kernel] {\label{fig:DMmv} \includegraphics[width=0.35\textwidth] {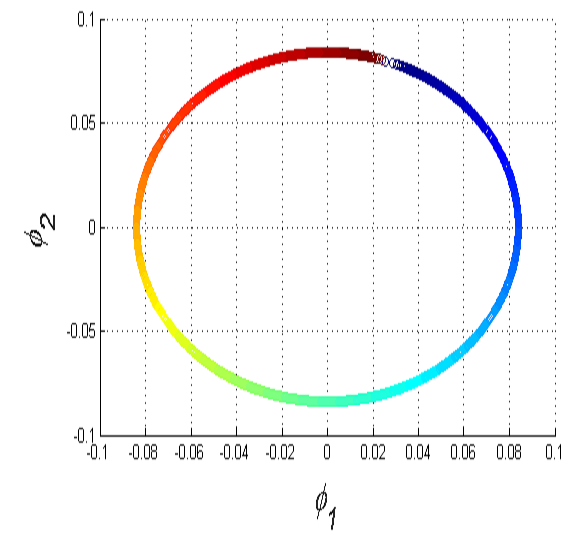}}
    \caption{Concatenation of DM coordinates $\bx_i~i=1,\ldots,20,000$ (a) compared to the DM application to the multi-view kernel approximated by Algorithm~\ref{alg:Mahalan} that considers the views $\btx_{i,l},~l=1,\ldots,10$ (b).}
    \label{fig:helixMVDM}
\end{figure}

%\begin{figure}[H]
%        \label{fig:helixSVDM}
%    \centering
%    \includegraphics[width=0.45\textwidth] {DM1.eps}
%    \includegraphics[width=0.45\textwidth] {DM2.eps}
%    \includegraphics[width=0.45\textwidth] {DM3.eps}
%    \caption{Two leading coordinated extracted using a single view DM.}
%\end{figure}
Next, the $10$ available views are concatenated to a single view $\bx_i = \left[\btx_{i,1},\ldots,\btx_{i,10} \right],~i=1,\ldots,2000$. The Mahalanobis distance is computed using Eq.~\ref{eq:mahal2} and DM is applied to the resulted kernel.  The first two leading coordinates of the kernel-based DM are presented in Fig.~\ref{fig:helixSVDM}. The large gap in each of the extracted manifolds is a deformation caused by the third coordinate of the transformation functions $\bbf_l (\btheta_i) = [{f_l^1}(\btheta_i),{f_l^2}(\btheta_i),{f_l^3}(\btheta_i)]$ (Eq.~\ref{eq: helixIII}). Furthermore, the output from the application of DM to the concatenation of the $10$ views is presented in Fig.~\ref{fig:DMconc}. The embedded manifold is even more distorted as the  gaps in the embedding suggests. Hence, the standard procedure, which concatenates the given set of features without considering the specific distortion each subset of features, may result in a distorted embedding.
%
%\begin{figure}[H]
%    \label{fig:helixMVDM}
%    \centering
%    \includegraphics[width=0.45\textwidth] {ConcatDM.eps}
%    \includegraphics[width=0.45\textwidth] {MultiDM.eps}
%    \caption{Left- extracted manifold using the concatenated data from all view (i.e kernel product), Right- Extracted manifold using the multi-view approach \ref{alg:Mahalan}.}
%\end{figure}
Finally, we apply Algorithm~\ref{alg:Mahalan} to all $10$ views and compute the two leading DM coordinates. The outputs are presented in Fig. \ref{fig:DMmv}. The algorithm overcomes deformations, seen as gaps in the concatenated embedding, by considering only the non-deformed small distances as the outcome of the majority vote function in Algorithm~\ref{alg:Mahalan}. The result is the circle shaped manifold that completely agrees with the corresponding intrinsic controlling angle parameter $\theta_i$.

\subsection{Classification of Seismic Events: Example III}
\label{sec:classification_of_sesmic_events}
To demonstrate the method in Section~\ref{sec:Multi-View of Dynamical Process} for analyzing a real data, we tackle the binary classification problem of seismic events. Discrimination between earthquakes and explosions is an essential component of nuclear test monitoring and it is also essential for creating reliable earthquake catalogs~\cite{rabin2016earthquake}. For this evaluation, we analyzed a dataset containing $46$ earthquakes and $62$ (non-nuclear) explosions recorded in Israel. The labels were annotated by a specialist from the Geophysical Institute of Israel (GII). 

\begin{figure}[H]
	\centering
	\includegraphics[width=0.9\textwidth] {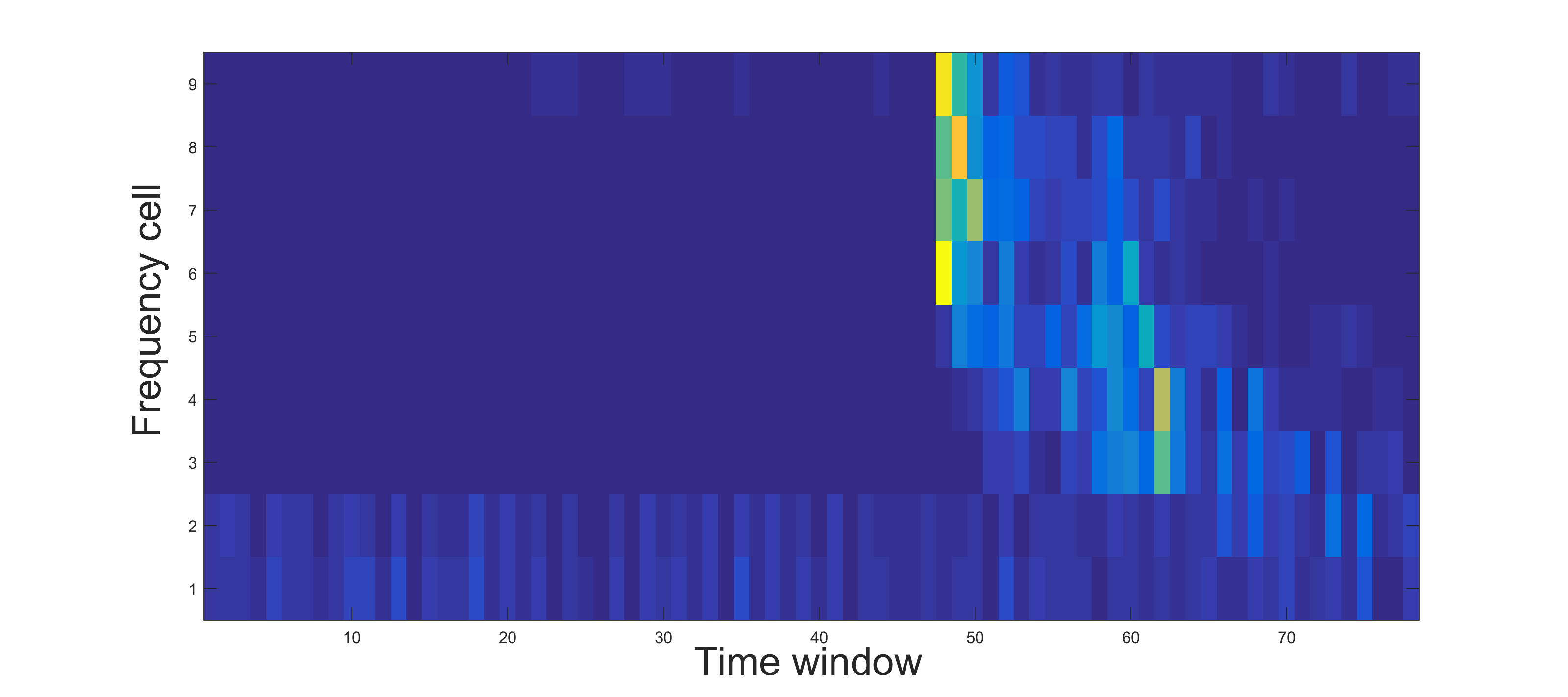}
	
	\caption{A Sonogram example of an explosion.}
	\label{fig:Sono}
\end{figure}

Seismic waveforms were recorded simultaneously using $3$ channels in two seismic monitoring stations. Each measurement channel is considered as a view to have a total of  $\zeta=6$ views. Since the seismic monitoring stations are far from each other, local noisy events such as constructions activities or vehicles movements are station specific. A Sonogram~\cite{Joswig} is computed from each measurement channel for each seismic event. The Sonogram is a time-frequency representation equally tempered on a logarithmic scale. Each Sonogram spectral component is normalized to have a maximum value of $1$. The detailed feature extraction process is given in \cite{rabin2016earthquake}. An example of a normalized Sonogram representation of an explosion is presented in Fig.~\ref{fig:Sono}. 
Under the assumption of Section~\ref{sec:Multi-View of Dynamical Process}, we consider each of the $6$ measurement channels as It\^{o} process. To evaluate Algorithm~\ref{alg:MVICA} in extracting the consensus from the given measurements, we added a Gaussian noise with zero mean and a variance $\sigma_N$ to $2$ views out of $6$. The added noise simulates an interference process that contaminates the two measurement channels. Since this noise is uncorrelated to the interesting seismic event it is not a part in the consensus. 

\begin{figure}[H]
	\centering
	\includegraphics[width=0.9\textwidth] {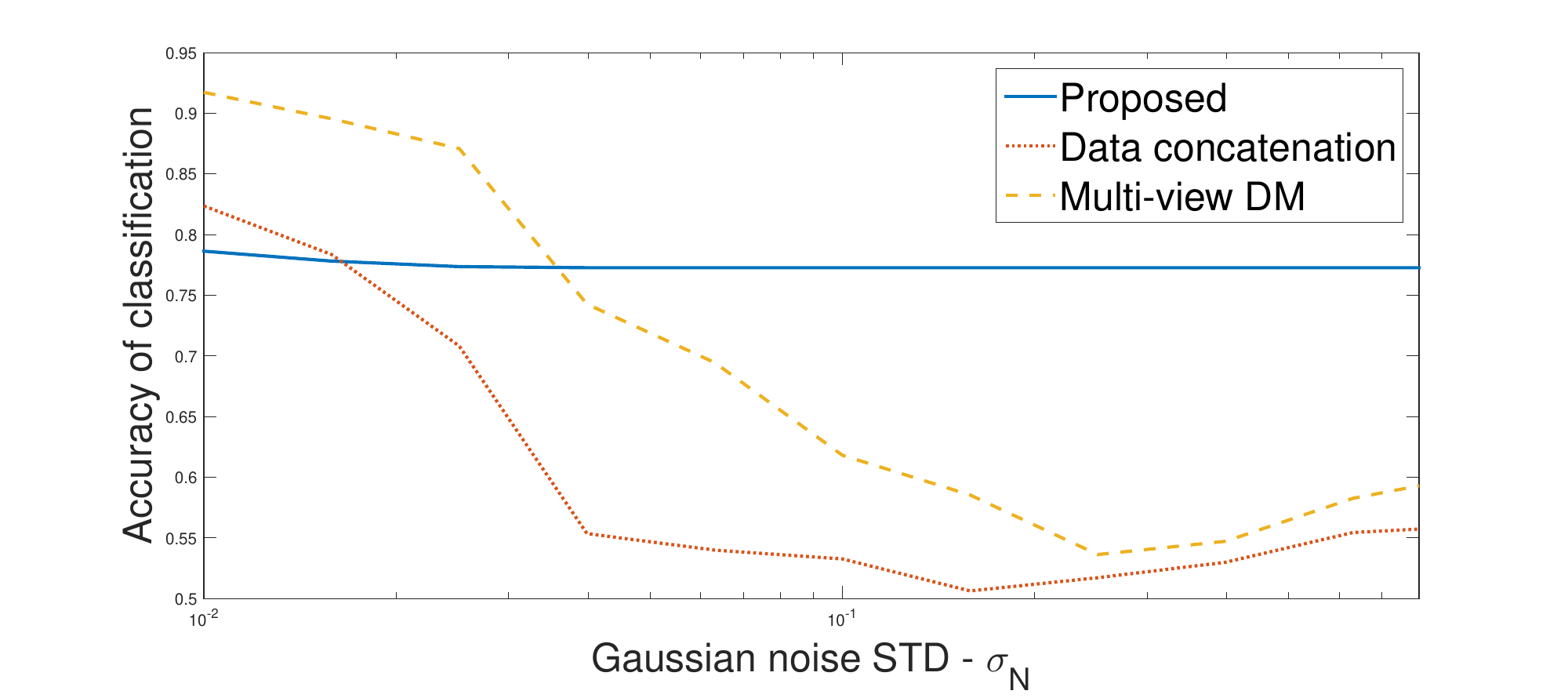}
	
	\caption{Average classification accuracy of seismic events by Algorithm~\ref{alg:MVICA} (blue), naive DM (dashed red) and multi-view-based DM~\cite{MV1}.}
	\label{fig:Sono2}
\end{figure}

Each feature vector $\tilde{\myvec{x}}_{i,l}$ represents the spectral properties of $i=1,...,78$ time window for  $l=1,..,6$ view. Algorithm~\ref{alg:MVICA} is applied followed by a spectral decomposition of $K_{\epsilon}$. The first nontrivial eigenvector is used as the representation for each seismic event.   
A binary classification is performed using K-NN ($K=5$) in a leave-one-out fashion. Furthermore, we applied the multi-view-based DM from~\cite{MV1} and a naive method of DM applied to a concatenation of the feature vectors $\tilde{\myvec{x}}_{i,l}, l=1,...,6$. The comparison between the resulted classification accuracies is presented in Fig.~\ref{fig:Sono2}. 

It is evident from Fig. \ref{fig:Sono2} that for measurements with a low power noise, the multi-view-based DM from~\cite{MV1} has the best accuracy of $91\%$. Application of Algorithm~\ref{alg:MVICA} results in a moderate accuracy of $79\%$. However, when  $\sigma_N$ is significant, Algorithm~\ref{alg:MVICA} overcomes the noise and is able to maintain a moderate accuracy while both the naive method and the multi-view-based DM  method are failing to less than $60\%$ accuracy. The performance of the proposed method is stable over large spans of noise power due to the availability of at least one reliable view for each classified seismic event.

\section{Conclusions}
\label{sec:Conclusions}

This paper presents a kernel construction scheme by approximating the similarity between intrinsic parameters that are common to multiple subset of features (also called views) in the presence of noise and non-linear transformation of the inaccessible controlling parameters. The presented method utilizes the relation between the Jacobian of each view and the corresponding Mahalanobis distance when a local covariance is approximated. This relation enables to approximate the affinities between intrinsic controlling parameters by considering the Mahalanobis distance of each view. The constructed kernel can be further normalized and decomposed to find an embedding of the data.
In order to demonstrate the effectiveness of the proposed method, we analyzed several synthetic datasets. This analysis showed that the correct affinities can be approximated in the presence of significant noise and a unknown non-linear transformation. Furthermore, in cases where the features are the output of a transformed intrinsic parameters with an associated non-full rank Jacobian, then the concatenation of the entire set of features results in a deformed manifold. In this case, the proposed multi-view scheme overcomes the problematic Jacobian  and  outputs a non-deformed manifold.
The proposed methodology involves a single spectral decomposition while increasing the number of (smaller Covariance-based) Mahalanobis distances computations per affinity. Hence, the growth in computation complexity is negligible.

\section*{Acknowledgment}
\noindent This research was partially supported by Indo-Israel CollAborative  for Infrastructure Security (Grant No. 1),
Israel Science Foundation (Grant No. 1556/17), 
US-Israel Binational Science Foundation (BSF 2012282),
Len  Blavatnik and the Blavatnik Family Foundation,  Blavatink ICRC Funds. 
The third author was partially supported by Fellowships from Jyv\"{a}skyl\"{a} University and the Clore Foundation.
We also would like to thank Aviv Rotbart for assisting with the experimental results.

%%%%%%%%%%%%%%%%%%%%%%%%%%%%%%%%%%%%%%%%%%%%%%%%%%%%%%%%%%%%%%%%%%%%%%%%%%
 \bibliographystyle{plain}
\bibliography{mv,listbMul}

\end{document}